\documentclass[11pt]{article}

\usepackage[T1]{fontenc}
\usepackage[utf8]{inputenc}
\usepackage{lmodern}
\usepackage[margin=1in]{geometry}
\usepackage{microtype}
\usepackage{amsmath,amssymb}
\usepackage{float}
\usepackage{tabularx}
\usepackage{array}
\usepackage{changepage}
\usepackage[numbers,sort&compress]{natbib}
\usepackage{url}
\usepackage{hyperref}
\usepackage{xcolor}
\usepackage{soul}
\newcolumntype{L}{>{\raggedright\arraybackslash}X}
\newcolumntype{C}{>{\centering\arraybackslash}X}
\newcolumntype{R}{>{\raggedleft\arraybackslash}X}
\newlength{\extralength}
\newcommand{\fulllength}{\dimexpr\textwidth+\extralength\relax}
\hypersetup{
    colorlinks=true,
    allcolors=blue,
    pdftitle={Spectrogram-Based CNN Autoencoder for One-Class Anomaly Detection in Machining Vibration Signals},
    pdfauthor={Sebastian Zeleny, Christian Gawron}
}

\usepackage{graphicx}
\usepackage{placeins}
\usepackage{wrapfig} 
\usepackage{multirow}
\usepackage{booktabs}
\usepackage{overpic}
\usepackage{hhline}
\definecolor{reviewerone}{RGB}{0,102,204} 
\definecolor{reviewertwo}{RGB}{213,94,0} 

	\title{Spectrogram-Based CNN Autoencoder for One-Class Anomaly Detection in Machining Vibration Signals}
	
	\author{
		Sebastian Zeleny\thanks{\raggedright Forst Technologie GmbH \& Co. KG, 42659 Solingen, Germany. Corresponding author: \href{mailto:zeleny@forst-online.de}{zeleny@forst-online.de}. ORCID: \href{https://orcid.org/0009-0002-5888-4892}{0009-0002-5888-4892}.}
		\and
		Christian Gawron\thanks{\raggedright FB Informatik \& Naturwissenschaften, Fachhochschule Südwestfalen, 58636 Iserlohn, Germany. \href{mailto:gawron.christian@fh-swf.de}{gawron.christian@fh-swf.de}. ORCID: \href{https://orcid.org/0000-0001-7139-7920}{0000-0001-7139-7920}.}
	}
	\date{}
	
\begin{document}
\maketitle

\begin{center}
\small Published in \textit{Applied Sciences} \textbf{16}(20), 10082 (2026).
DOI: \href{https://doi.org/10.3390/app162010082}{10.3390/app162010082}.
\end{center}

\begin{abstract}
This work presents a {one-class anomaly} detection framework for vibration-based condition monitoring in machining processes using a 
	convolutional neural network autoencoder. Three-axis vibration signals acquired during industrial broaching are transformed into time--frequency 
	spectrograms and combined into a compact three-channel representation, enabling convolutional feature learning on multivariate time-series data. 
	The autoencoder is trained exclusively on normal process data and detects anomalous strokes based on reconstruction errors. 
	{Labeled validation data are used for threshold calibration and an independent test set is reserved for final performance evaluation.}
	To improve sensitivity to short-lived process instabilities such as chatter, a localized residual-based metric termed Segmented Vertical Error (SVE) 
	is introduced. Experimental results on approximately 10{,}000 broaching strokes show a clear separation between normal and anomalous process conditions 
	under strong class imbalance. The processing pipeline is additionally evaluated on an external public turning dataset as a proof of concept, indicating 
	preliminary applicability beyond the original broaching setup. The results suggest that spectrogram-based autoencoder models can provide a practical and 
	scalable basis for one-class anomaly detection in industrial vibration monitoring.
\end{abstract}

	

		\section{Introduction}
		
		Condition monitoring is a key element of modern maintenance strategies in manufacturing, enabling the early detection of abnormal process conditions 
		and machine degradation before severe failures occur \cite{WKZ_Maschinen}. By identifying deviations from stable operating behavior, 
		condition monitoring contributes to improved product quality and increased process reliability, particularly in highly automated production environments.
		
		Among the various sensor modalities used for machine condition monitoring, vibration signals are widely regarded as one of the most informative indicators for machining processes \cite{CondVibration, CondMonitor}. Vibrations directly reflect the dynamic interactions between tool, workpiece, and machine structure, and even minor changes in the mechanical state can lead to measurable variations in the vibration signature. 
		Numerous studies have shown that vibration analysis enables the early detection of tool wear, misalignment, bearing defects, and unstable cutting conditions \cite{PraxisSchwingung, SensorStudy, VibrationStudy}.
		
		Broaching is a highly productive but dynamically sensitive machining process, characterized by periodically varying cutting forces and the symmetric arrangement of cutting teeth \cite{Forst, Fertigungsverfahren}. These characteristics can induce coupled vibrations between tool and workpiece, which may result in chatter, surface defects, and form deviations \cite{AXINTE_CoupledVibration}. 	
		Due to the typically automated and largely unmonitored operation of broaching machines, reliable process and condition monitoring is of particular importance \cite{Axinte2007}. 
		Existing research on broaching process monitoring has predominantly focused on the analysis of force–time diagrams as well as on the combined evaluation of force, vibration, and acoustic emission signals, often relying on handcrafted features and expert knowledge \cite{AXINTE_Process_Monitoring}.
		While these approaches have demonstrated their capability to identify process anomalies and tool wear, they often require process-specific feature engineering and expert-defined thresholds, limiting their robustness and transferability across different machines and operating conditions \cite{MultiSensorBroaching,ToolWearMonitoring,ALDEKOA2023}. 
				
		Similar challenges arise in other machining processes such as milling, drilling, and turning, 
		where tool, workpiece, and machine vibrations likewise contribute to surface defects and production 
		errors \cite{ARRAZOLA2020,ARRAZOLA2021,ARRAZOLA2023b}.
		This indicates that vibration-induced anomalies are not specific to a single process, but represent 
		a general challenge in machining operations.
	
		In industrial practice, comprehensive fault labels are rarely available, whereas large amounts of 
		data from stable operating conditions can typically be collected. Consequently, {one-class anomaly 
		detection} is a suitable approach for vibration-based condition monitoring \cite{Autoencoders}.
		
		Although existing unsupervised approaches commonly operate directly on vibration time series or 
		handcrafted features, they often struggle to capture complex temporal–spectral patterns and 
		correlations across multiple sensor channels \cite{Chen2021}.
		To address these limitations, the proposed approach transforms multi-axis vibration signals into 
		spectrogram-based image representations and applies convolutional autoencoder models for anomaly 
		detection \cite{UNet, Gramelt2024}.

		The main contributions of this work are as follows: 
		(i) a spectrogram-based representation for three-axis broaching vibration signals, 
		(ii) a quantitative evaluation on industrial broaching data containing rare process anomalies, 
		(iii) a localized residual metric termed Segmented Vertical Error (SVE) for detecting short-lived vibration anomalies, and 
		(iv)~an additional {cross-process proof of concept} on a public turning dataset to assess the applicability of the pipeline to another machining process.
		
		\section{Materials and Methods}
		\subsection{Data Description}
		
\begin{figure}[H]			
			\includegraphics[width=0.4\textwidth]{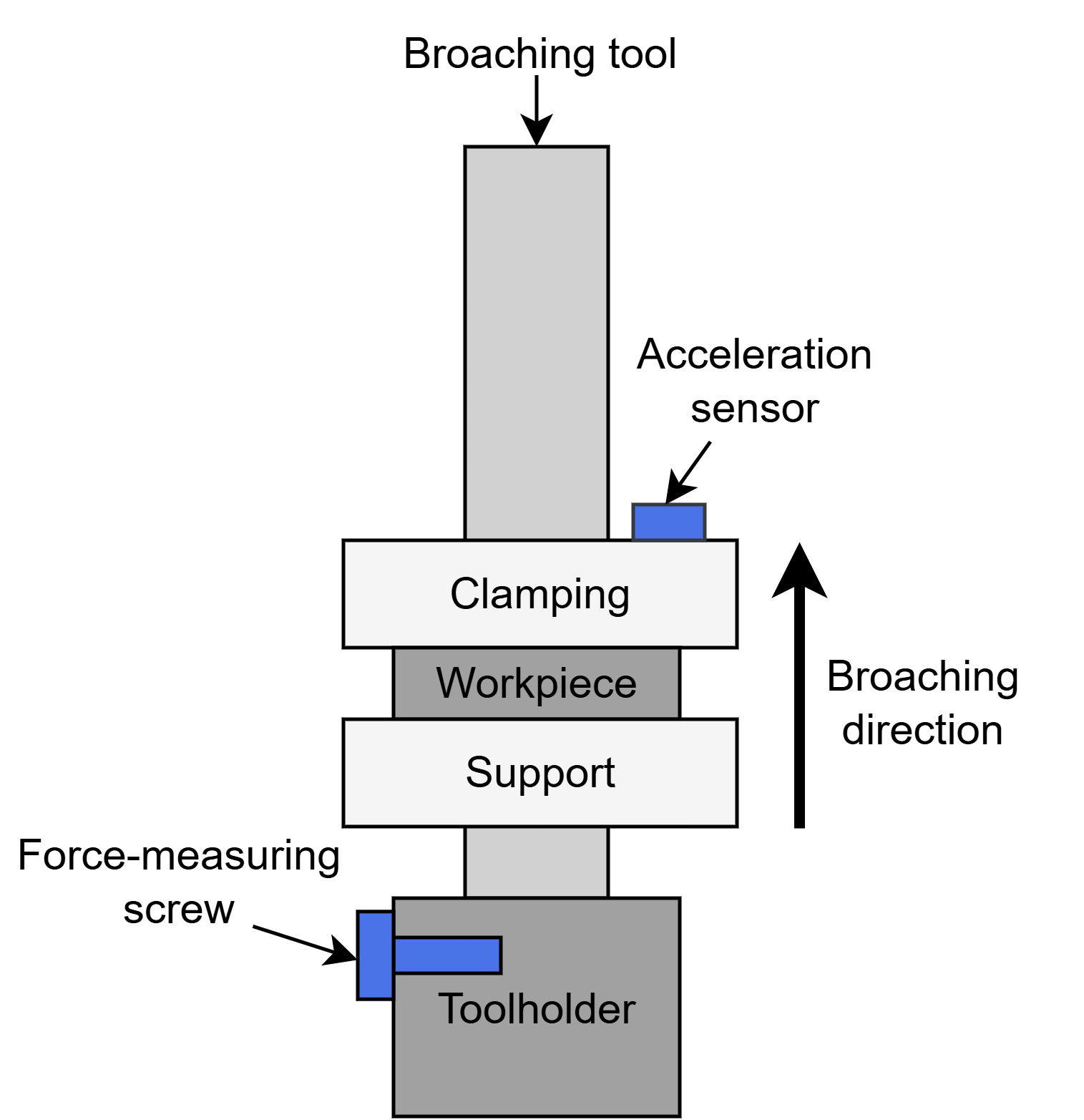}
			\caption{Schematic
 illustration of the experimental measurement setup and sensor placement.}
			\label{fig:schemeprototype}
		\end{figure}

		The experimental data were acquired from an industrial broaching machine used in the production 
		of internal gears (ring gears). The broaching process involves a vertical feed motion of the 
		broaching tool—consisting of a base body and a bushing—into the workpiece to generate the desired tooth profile. 
		A three-axis accelerometer was mounted in close proximity to the workpiece clamping unit in order to capture 
		vibration signals of the clamping plate along all three spatial directions. The sensor signals were sampled at 2.5~kHz.
		The experimental setup and sensor placement are shown in Figure~\ref{fig:schemeprototype}.
		 
		{The measurement setup was designed to capture chatter-related vibrations and other process instabilities in close 
		proximity to the machining zone. During the initial characterization of these process instabilities, force and three-axis 
		acceleration signals were acquired synchronously. A force-measuring screw integrated into the toolholder assembly was used 
		to capture the force response during machining. Dynamic force fluctuations associated with pronounced vibration events could 
		therefore be observed in the corresponding force--time signals.} 

		{The corresponding acceleration time series and spectrograms were then examined to determine how these events 
		manifest in the vibration-based representation. 
		These representative comparisons included vibration events of different 
		intensities and events occurring during different phases of the broaching stroke, including roughing and finishing. 
		They provided the physical reference for identifying characteristic vibration patterns in the spectrogram representation. 
		The force measurements were used for this initial characterization and were not used for model training.}
		
		The complete dataset comprises approximately 10{,}000 individual time series, each corresponding 
		to a single broaching stroke and acquired using a start-triggered measurement scheme. 
		Due to confidentiality restrictions, raw time-domain signals cannot be disclosed; however, 
		the subsequent signal processing and feature extraction pipeline is described in detail.
		
		Approximately 1\% of the recorded strokes exhibit anomalous vibration behavior, resulting in a 
		highly imbalanced dataset. Following a {one-class learning strategy}, model training was 
		performed exclusively on data representing stable operating conditions. 

		{For evaluation purposes, the validation and test data were retrospectively reviewed using the characteristic 
		spectrogram patterns identified during the initial force--vibration comparisons described above. 
		Each stroke was manually annotated as either normal or anomalous based on the presence of these characteristic patterns in the 
		spectrogram representation.
		All evaluation spectrograms were annotated by one of the authors prior to model evaluation. 
		Consequently, model predictions and reconstruction errors were not used during the annotation process.}

		{Not every individual spectrogram annotation was independently verified against its corresponding force signal. The resulting labels are therefore considered force-informed expert annotations rather than 
		independently force-verified ground truth.}
		
		To assess whether the proposed processing pipeline is applicable beyond the original broaching 
		setup, a publicly available vibration dataset from a turning process was additionally {evaluated as a cross-process proof of concept} 
		and is described in Section~\ref{validation}.
		
		\subsection{Qualitative Description of Process Anomalies}

\begin{figure}[H]
			 
			\includegraphics[width=\linewidth]{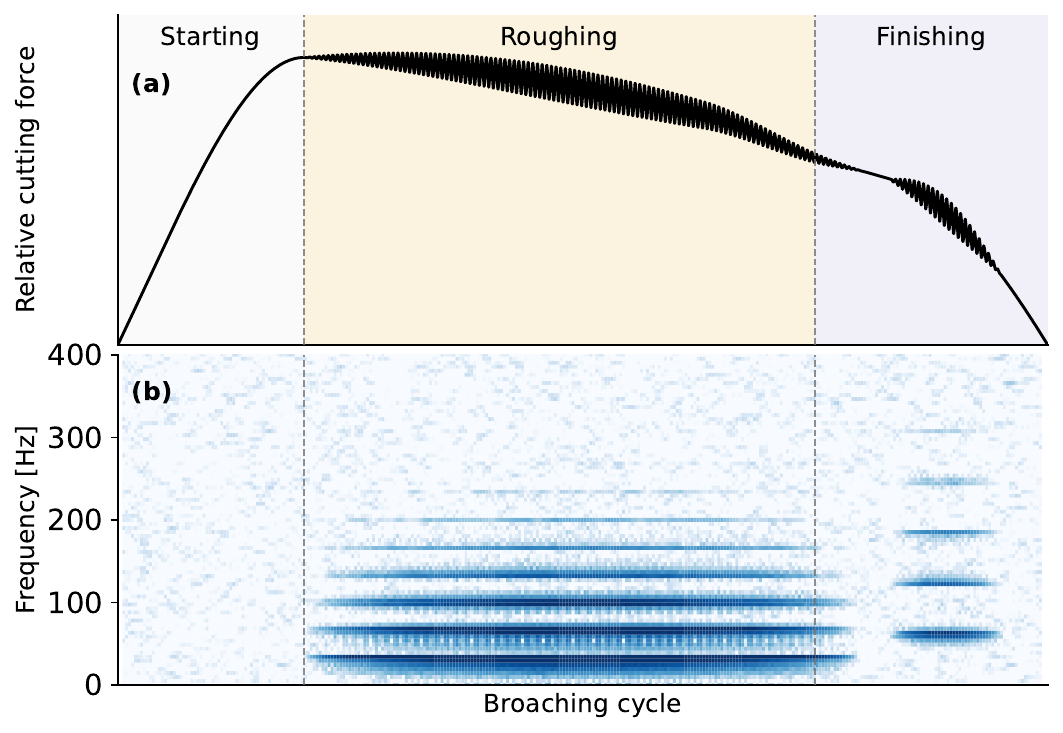} 
			\caption{Schematic 
 relationship between oscillatory fluctuations in the cutting-force
			signal (\textbf{a}) and corresponding vibration signatures in the acceleration spectrogram (\textbf{b}) during a broaching stroke.} 
			\label{fig:stroke_representations} 
		\end{figure}

		Figure~\ref{fig:stroke_representations} schematically illustrates the relationship underlying the annotation procedure. 
		In the force--time signal (Figure~\ref{fig:stroke_representations}a), vibration-related process instabilities appear as pronounced 
		oscillatory fluctuations of different intensities and may occur during different phases of the broaching stroke. In the corresponding 
		acceleration spectrogram (Figure~\ref{fig:stroke_representations}b), these transient events appear as localized time--frequency structures. 
		Representative comparisons of synchronous force and acceleration measurements were used to characterize these vibration patterns and 
		provided the physical reference for the spectrogram patterns subsequently used during expert annotation. Since force measurement systems 
		are often difficult to integrate into industrial machines, acceleration sensors provide a less intrusive alternative for vibration-based 
		process monitoring. Vibrations during the finishing phase are particularly relevant because disturbances occurring during this stage may 
		affect the final workpiece surface.

		\subsection{Signal Processing and Image-Based Representation}
		\label{signalProcessing}
		
		To effectively capture multivariate vibration behavior during broaching, the three-axis acceleration signals (X, Y, Z) are transformed into time–frequency representations using the Short-Time Fourier Transform (STFT). 
		The STFT is computed using fixed parameters for all signals to ensure consistent representation, as summarized in Table~\ref{tab:stft_params}. 

		\begin{table}[H] 
			 
			\caption{Parameters used for STFT-based spectrogram generation.} 
			\label{tab:stft_params} 
			\begin{tabularx}{\textwidth}{LL} \toprule \textbf{Parameter} & \textbf{Value} \\ \midrule Sampling rate & \(2.5\,\mathrm{kHz}\) \\ Window length & \(512\) 
				samples \\ Window function & Hann \\ Overlap & \(75\,\%\) \\ Frequency range & \(0\text{--}600\,\mathrm{Hz}\) 
				\\ Output resolution & \(150 \times 100\) pixels \\ 
				\bottomrule 
			\end{tabularx} 
		\end{table}

		Power spectral densities were converted to decibel scale, clipped to the range [$-$58 dB, 0 dB], normalized to [0, 1], and combined into a three-channel RGB representation.
		Each axis is converted into a spectrogram, preserving temporal, spectral, and amplitude-related characteristics of the vibration signals.

\begin{figure}[H]
			
			\hspace{-0.4cm}\includegraphics[width=1.0\textwidth]{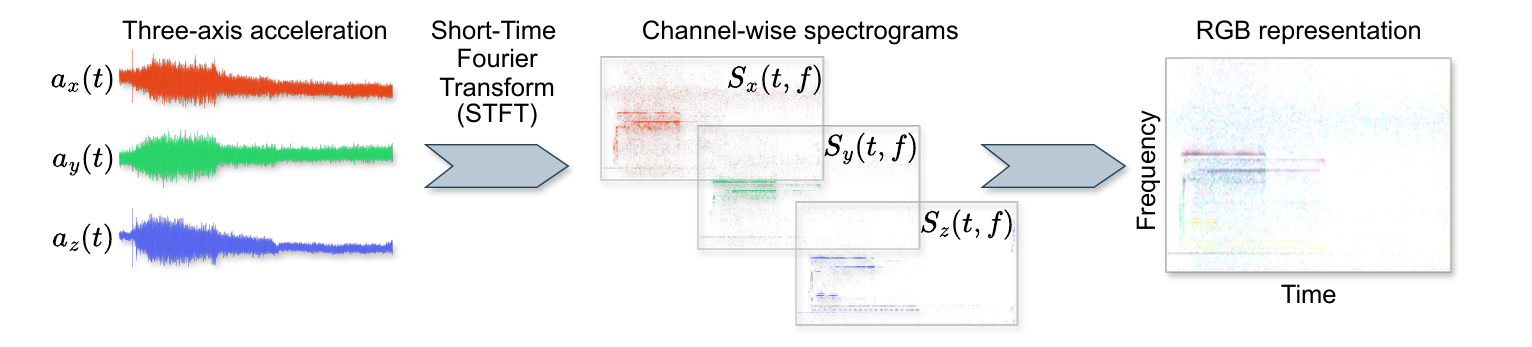}
			\caption{Transformation
 pipeline of three-axis acceleration signals into a single RGB time--frequency 
			representation using the short-time Fourier transform (STFT). Each axis is represented in a separate color channel (X = red, Y = green, Z = blue).}
			\label{fig:pipeline_rgb}
		\end{figure}

		To enable CNN-based anomaly detection, the three spectrograms were combined into a single 
		false-color image: the X-axis is represented in red, the Y-axis in green, and the Z-axis in blue. 
		This approach maps the multivariate time series into a compact image representation, allowing correlations between axes, 
		frequency content, and temporal evolution to be simultaneously encoded, as illustrated in Figure~\ref{fig:pipeline_rgb}.
		
		Beyond its conceptual advantages, the proposed image-based representation significantly reduces data dimensionality. 
		Raw multivariate time-series data typically consist of high-resolution signals sampled at high frequencies, resulting 
		in a large number of samples per channel and high memory requirements for storage and processing. 
		This compression enables more efficient storage and faster processing, making the approach well suited for large-scale 
		industrial applications and real-time inference.

\begin{figure}[H]
			
			\fcolorbox{gray!50}{white}{
				\includegraphics[width=0.46\textwidth]
				{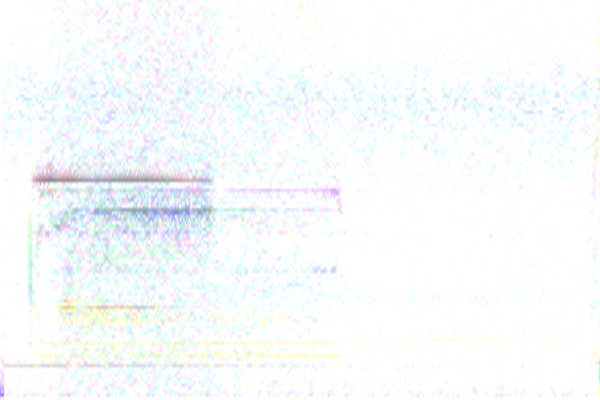}
			}
			\hfill
			\fcolorbox{gray!50}{white}{
				\includegraphics[width=0.46\textwidth]
				{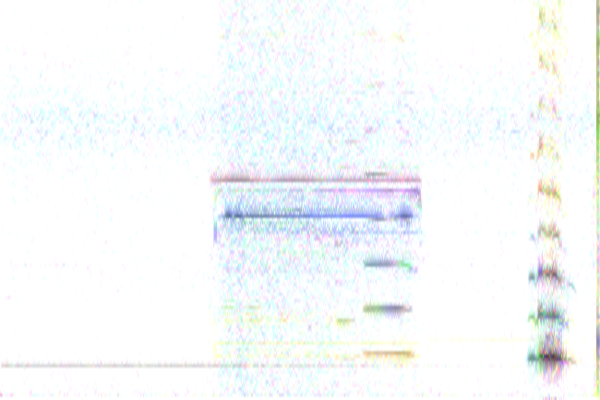}
			}
		\caption{Spectrogram
 image of three-axis vibration data: (\textbf{left}) normal broaching stroke and (\textbf{right}) stroke containing 
			anomalous vibration behavior. Each color channel encodes one sensor axis (X = red, Y = green, Z = blue).}
			\label{fig:spectrogram_rgb}
		\end{figure}

		Figure~\ref{fig:spectrogram_rgb} shows example spectrograms for normal and anomalous broaching strokes. 
		Anomalies appear as localized high-energy patterns that deviate from the characteristic structure observed during stable 
		operation. These spectrograms serve as input to the anomaly-detection models described in the following sections.
		
		The resulting RGB spectrograms serve as input to the convolutional autoencoder used for anomaly detection across 
		multiple sensor channels.

		\subsection{CNN Autoencoder Architecture}
		
\begin{figure}[H]
			
			\includegraphics[width=0.95\textwidth]{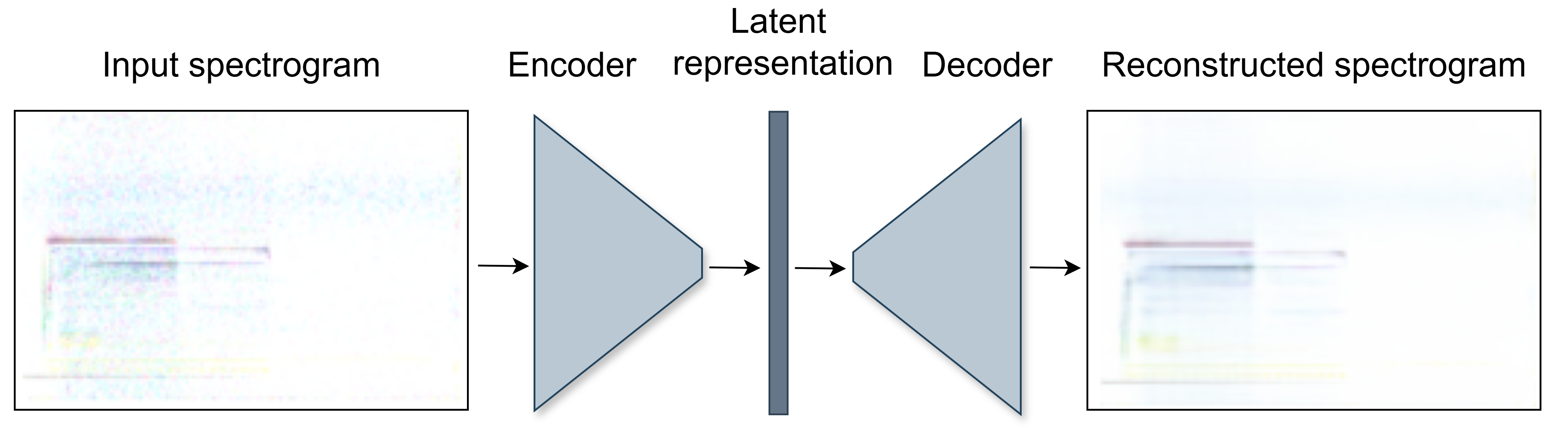}
			\caption{Conceptual
 overview of the convolutional autoencoder.}
			\label{fig:ae}
		\end{figure}

		Building on the spectrograms introduced in the previous section, this section describes the convolutional autoencoder 
		(CNN-AE) architecture used for learning compact representations of normal process behavior and detecting anomalies via 
		reconstruction errors. Convolutional autoencoders combine local convolutional feature extraction with an encoder--decoder 
		reconstruction objective and have been established for unsupervised hierarchical feature learning in image-like data \cite{Masci2011CAE}.
		The overall architecture is shown in Figure~\ref{fig:ae}.

		The input to the CNN autoencoder consists of normalized spectrogram images with a width of 150 pixels and a height of 100 pixels. Each image has three color channels, where each channel represents the time–frequency information of one acceleration axis.
		
		The encoder consists of multiple convolutional layers with increasing filter depth and nonlinear activation functions. These layers progressively reduce the spatial resolution of the input using pooling operations and encode characteristic vibration patterns into a low-dimensional latent representation.
						
		The bottleneck represents the compressed latent space of the model and is deliberately constrained to capture only the 
		essential characteristics of normal operating conditions. This limitation is critical for anomaly detection, as it prevents 
		the network from accurately reconstructing previously unseen patterns.
		
		The decoder  mirrors the encoder architecture and reconstructs the input image from the latent representation using upsampling
		and convolutional layers.

		The architecture shown in Table~\ref{tab:autoencoder} served as the baseline model for the subsequent experiments. 
		Hyperparameter studies were conducted to evaluate the influence of bottleneck size, learning rate, batch size, and training duration \cite{BottleneckInvestigation, Hyper-Parameter}.

				\begin{table}[H] 
			 
			\caption{Architecture of the convolutional autoencoder. Output dimensions are given as height $\times$ width $\times$ channels.} 
			\label{tab:autoencoder} 
			\begin{tabularx}{\textwidth}{LLL} \toprule \textbf{Stage} & \textbf{Layer Configuration} & \textbf{Output Size} \\ \midrule Input & RGB spectrogram image & \(100 \times 150 \times 3\) \\ 
				\addlinespace Encoder & Conv (4) $\rightarrow$ Pool (\(2 \times 2\)) & \(50 \times 75 \times 4\) \\ & Conv (8) $\rightarrow$ Pool (\(2 \times 3\)) 
				& \(25 \times 25 \times 8\) \\ & Conv (12) & \(25 \times 25 \times 12\) \\ \addlinespace Bottleneck & Dense & \(16\) \\ 
				\addlinespace Decoder & Dense $\rightarrow$ Reshape & \(25 \times 25 \times 12\) \\ & Conv (12) $\rightarrow$ Up (\(2 \times 3\)) 
				& \(50 \times 75 \times 12\) \\ & Conv (8) $\rightarrow$ Up (\(2 \times 2\)) & \(100 \times 150 \times 8\) \\ & Conv (4) & \(100 \times 150 \times 4\) \\ & Conv (3, sigmoid) & \(100 \times 150 \times 3\) \\ 
				\bottomrule 
			\end{tabularx} 
		\end{table}

		\subsection{Classical Baseline Methods}
		\label{baseline}
		
		{To contextualize the autoencoder-based results, three classical image-based anomaly detection baselines are 
		evaluated on the evaluation datasets. 
		To ensure a controlled comparison with the CNN autoencoder, all baseline methods retain the same three-axis 
		information encoded by the RGB spectrogram representation.}
		The primary broaching evaluation uses the frozen temporal validation/test split: models are fitted only on 
		nominal training images, decision thresholds are selected on labeled validation data by maximizing the F1-score, 
		and the selected thresholds are then kept fixed for the independent test evaluation.
		
		The first two baselines operate on handcrafted image descriptors {derived separately from all three spectrogram 
		channels (X/Y/Z)}, including summary intensity statistics, gradient statistics, intensity histograms, and 
		row- and column-wise aggregate profiles. {The channel-wise descriptors are concatenated} and standardized 
		using statistics from the nominal training data. A one-class support vector machine with an RBF kernel, 
		$\nu=0.05$, and \texttt{gamma=scale}, and an isolation forest with \texttt{contamination=auto} are then 
		fitted to the standardized nominal descriptors. Their negative decision-function values are used as anomaly 
		scores; the isolation forest uses random seed 42.
		
		The third baseline uses principal component analysis (PCA) as a linear reconstruction model. 
		{The complete three-channel RGB spectrograms} are flattened into image vectors, and a PCA model with 
		up to 32 components and the randomized solver is fitted on nominal training images. The mean squared 
		reconstruction error is used as the anomaly score.

		\subsection{Training Strategy and Anomaly Detection}

		The convolutional autoencoder is trained exclusively on spectrograms representing normal operating conditions. 
		This one-class training strategy reflects typical industrial scenarios in which labeled fault data are scarce or unavailable.
		{Accordingly, the term one-class refers to the training of the autoencoder on nominal data only.}

		Prior to training, all spectrogram images are normalized to the range $[0,1]$. 
		The network is optimized using the Adam optimizer, and the mean squared reconstruction error (MSE) 
		between the input image $x_i$ and the reconstructed output $\tilde{x}_i$ is minimized:
		\begin{equation}
			\mathcal{L}_{\text{MSE}} = \frac{1}{N} \sum_{i=1}^{N} \left\| x_i - \tilde{x}_i \right\|^2 .
		\end{equation}
		
		Although a low reconstruction error indicates accurate image reconstruction, it does not necessarily 
		correspond to optimal anomaly detection performance. Therefore, the training process is evaluated 
		using detection-oriented metrics, such as the F1-score and the precision–recall area under the 
		curve (PR-AUC), computed on a validation subset.
		
		During inference, pixel-wise reconstruction errors are aggregated into an anomaly score. 
		{For the primary broaching evaluation, the decision threshold is selected on the validation set by 
		maximizing the F1-score and subsequently applied unchanged to the independent test set. 
		The separate run-wise calibration protocol used for the turning dataset is described in 
		Section~\ref{validation}.}
		
		The final model configuration was selected based on systematic hyperparameter studies. 
		The results indicate that a moderate bottleneck size provides the best balance between 
		sensitivity and generalization. The final architecture employs a bottleneck dimension of 16, 
		a batch size of 32, and a learning rate of $1 \times 10^{-4}$. This configuration achieved
		the highest overall detection performance while maintaining stable training behavior.
		The complete layer-wise specification and reproducibility settings are provided in
		Appendix~\ref{app:ae_architecture}.
		
		Due to the highly imbalanced dataset, where anomalies account for less than 1\% of all samples, 
		PR-AUC is used as the primary evaluation metric. Unlike ROC-AUC, which may provide overly 
		optimistic estimates under severe class imbalance, PR-AUC focuses directly on precision and 
		recall and is therefore more informative for rare-event detection tasks \cite{PR-AUC,ROC,ROCAUC}.
		Consequently, PR-AUC is used as the primary evaluation metric throughout this study.
				
		\subsection{Dataset Construction and Preprocessing} 
		\label{Dataset}

		The dataset originates from a single industrial broaching machine and was divided into temporally 
		separated subsets. Due to tool changes, wear progression, and process variability, the resulting 
		subsets exhibit shifted statistical properties.

		A subset of 5000 samples was selected from an earlier period representing stable operating 
		conditions and used exclusively for model training. 
		In accordance with the one-class learning paradigm, the training set contains only samples 
		corresponding to normal process behavior.

		A second subset of 5000 samples was selected from a later period, during which varying operating 
		conditions, tool changes, and increased process variability occurred. 
		This dataset includes both normal and anomalous samples and is further divided into validation 
		and test sets.

		The validation set is used for hyperparameter tuning and threshold selection, whereas the test 
		set remains strictly held out for final performance evaluation.
		{This temporally separated training/validation/test configuration constitutes the primary evaluation 
		protocol for the industrial broaching dataset.}

\begin{figure}[H]
			\includegraphics[width=1\textwidth]{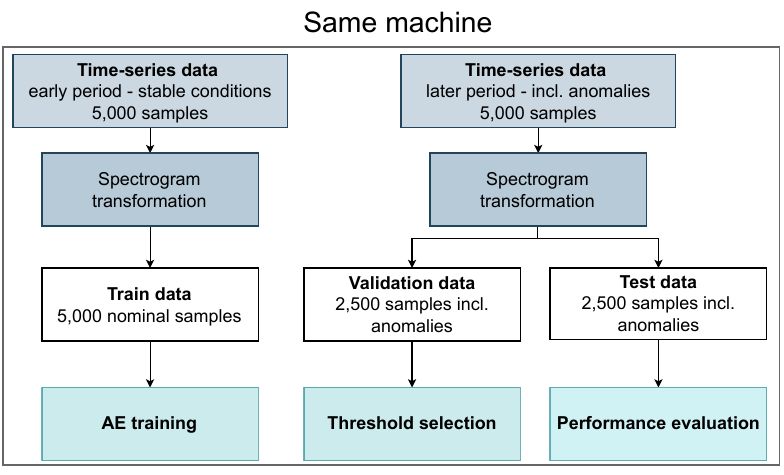}
			\caption{Primary
 evaluation protocol for the industrial broaching dataset. The temporally separated data 
			are divided into nominal training data, labeled validation data for decision-threshold calibration, 
			and an independent test set for final performance evaluation.}
			\label{fig:dataset}
		\end{figure}

		Figure~\ref{fig:dataset} summarizes the dataset construction and splitting procedure.

\vspace{-4pt}
		
		The final training set thus represents a consistent approximation of normal process behavior, 
		while the independent test set remains unfiltered and contains both normal and anomalous samples. 
		This strict separation ensures unbiased performance evaluation on the industrial test data.
		
		The setup reflects practical industrial scenarios, where models trained on historical data are 
		deployed on future data subject to process drift and evolving system states.
						
		\section{Results}
			
		This section presents the experimental results of the proposed spectrogram-based anomaly detection framework. 
		The evaluation focuses on reconstruction behavior, separation of normal and anomalous samples, and quantitative detection 
		performance measured using the area under the Precision–Recall curve (PR-AUC).
		
		\subsection{Reconstruction}
		
\begin{figure}[H]
			
			\renewcommand{\arraystretch}{1.3}

			\hspace{-0.2cm}\begin{tabular}{c c c c}

				& \textbf{Input}
				& \textbf{Reconstruction}
				& \textbf{Residual} \\[0.2cm]

				\textbf{Nominal}
				& \fcolorbox{gray!50}{white}{
					\includegraphics[width=0.22\textwidth]{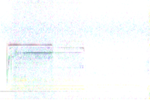}
				}
				& \fcolorbox{gray!50}{white}{
					\includegraphics[width=0.22\textwidth]{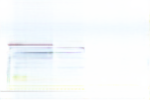}
				}
				& \fcolorbox{gray!50}{white}{
					\includegraphics[width=0.22\textwidth]{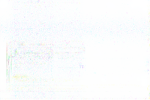}
				} \\[0.4cm]

				\textbf{Anomaly}
				& \fcolorbox{gray!50}{white}{
					\includegraphics[width=0.22\textwidth]{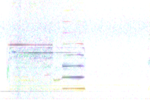}
				}
				& \fcolorbox{gray!50}{white}{
					\includegraphics[width=0.22\textwidth]{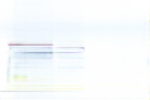}
				}
				& \fcolorbox{gray!50}{white}{
					\includegraphics[width=0.22\textwidth]{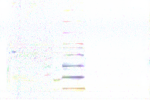}
				}

			\end{tabular}

			\caption{Qualitative
 reconstruction results for normal and anomalous samples. 
			Normal spectrograms produce only minor residuals, whereas anomalous samples result in 
			structured reconstruction errors.
			}

			\label{fig:reconstruction_examples}
		\end{figure}

		To qualitatively assess the behavior of the trained autoencoder, Figure~\ref{fig:reconstruction_examples} shows representative normal and anomalous spectrograms, their reconstructions, and the corresponding residuals.
		
\vspace{-4pt}

		For normal process conditions, the reconstructed images closely resemble the corresponding inputs and preserve the dominant time–frequency structures. 
		The resulting residual maps exhibit only minor, spatially unstructured deviations, indicating that the learned latent representation accurately captures characteristic process dynamics.
		
		In contrast, spectrograms containing anomalous vibration patterns show clearly visible reconstruction discrepancies. 
		These deviations are particularly pronounced in frequency regions associated with irregular oscillatory components and lead to structured residual patterns.
	
		\subsubsection*{Segmented Vertical Error (SVE)} 
		
		In many reconstruction-based approaches, anomaly detection is performed using the global 
		reconstruction error. 
		However, relevant process anomalies often appear as localized vertical structures in spectrograms, 
		which may be diluted when the error is averaged over the entire image.

		{In the spectrogram representation used here, the horizontal axis represents time and the vertical axis represents frequency. 
		Short-lived machining instabilities are therefore confined to a limited region along the time axis, while the associated vibration 
		response may simultaneously affect multiple frequency components. Such transient events can consequently appear as vertically 
		extended, temporally localized structures in the \mbox{time--frequency} representation. Short-time spectral analysis has accordingly 
		been used to identify and track chatter-related vibration components in machining processes} \cite{Spectrogram-Chatter}.
				
		To address this limitation, a localized error metric referred to as the Segmented Vertical Error (SVE)
		is introduced. 
		The residual image, obtained from the difference between input and reconstructed spectrogram, is divided into multiple equally sized vertical segments. 

		For each vertical segment $S_j$, a local reconstruction error 
		\begin{equation} 
			e_j= \frac{1}{|S_j|} \sum_{p \in S_j} |x_p-\tilde{x}_p| 
		\end{equation} 
		is computed, where $N_{\mathrm{seg}}$ denotes the number of vertical segments, $S_j$ the set of pixels 
		belonging to the $j$-th segment, $|S_j|$ the number of pixels in that segment, and $x_p$ and 
		$\tilde{x}_p$ the original and reconstructed pixel values, respectively. 
		The final anomaly score is then defined as 
		\begin{equation} 
			\mathrm{SVE} = \max_{1 \le j \le N_{\mathrm{seg}}} e_j. 
		\end{equation}

\begin{figure}[H]
		\begin{tabular}{@{}cc@{}} \text{Nominal sample} & \text{Anomalous sample} \\[2pt] 
			\includegraphics[width=0.32\textwidth]{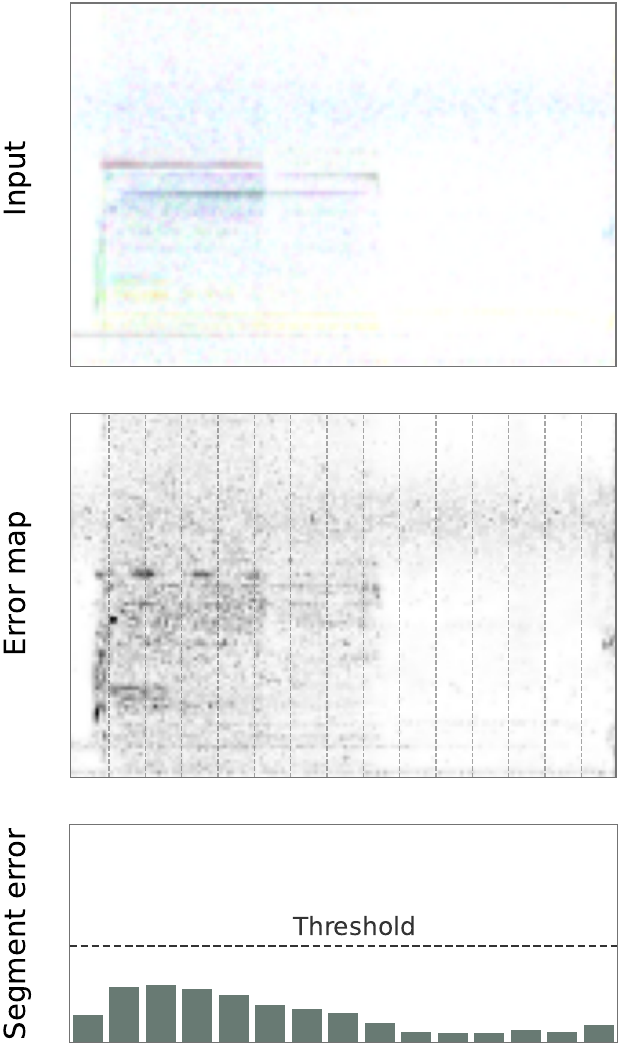} & 
			\includegraphics[width=0.32\textwidth]{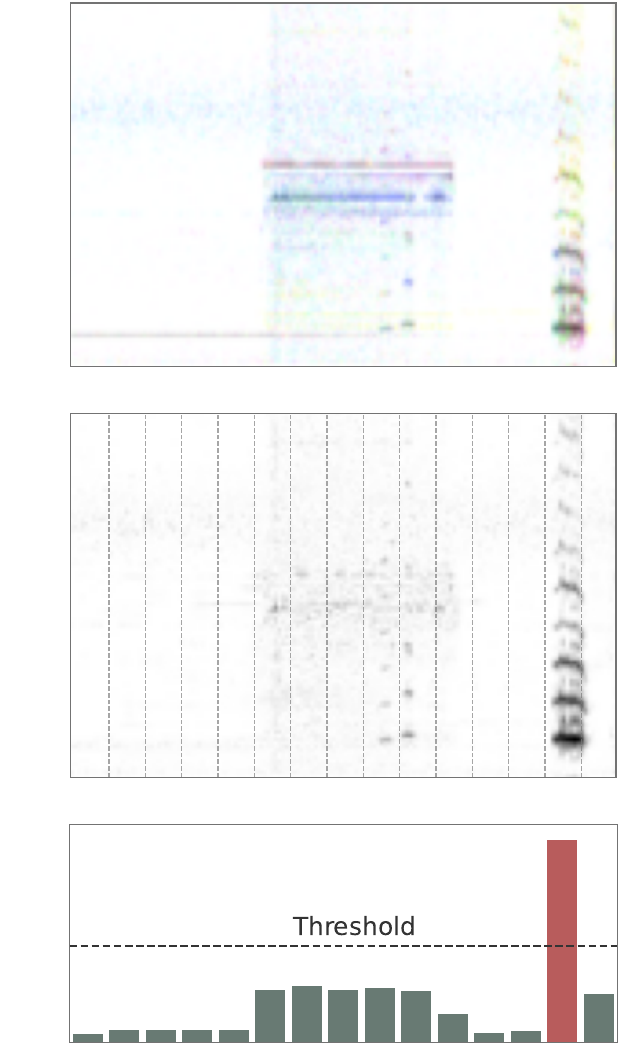} 
		\end{tabular}
			\caption{Segment-wise
 anomaly detection for a nominal and an anomalous sample. Reconstruction 
			errors are aggregated over vertical spectrogram segments and compared with the decision threshold; 
			threshold exceedance indicates anomalous behavior.} 
			\label{fig:sve} 
		\end{figure}

		This maximum operation prevents short localized deviations from being diluted by averaging the 
		reconstruction error over the entire spectrogram.
		Figure~\ref{fig:sve} illustrates the segment-wise error calculation for nominal and anomalous samples.
		
		This segmentation enables the detection of localized anomaly patterns that manifest as vertically 
		concentrated reconstruction errors. In contrast, reconstruction errors originating from 
		horizontally distributed structures are spread across multiple segments and therefore have less 
		influence on the final anomaly score.

		An additional advantage of the approach is that segments with elevated error values can be 
		directly associated with specific regions of the spectrogram, providing a coarse localization of 
		anomalous process behavior. Furthermore, predefined regions, such as boundary areas or known 
		non-critical sections, can easily be excluded from the evaluation if required.

		The resulting segment-wise error values are subsequently evaluated using the same thresholding 
		strategy described previously, enabling anomaly detection based on localized reconstruction 
		deviations rather than a purely global reconstruction error.
				
		\subsection{Detection Performance on the Broaching Dataset}

		To evaluate the detection performance of the proposed method, the anomaly score threshold was 
		systematically varied over the full score range. 
		For each threshold, precision and recall were computed to obtain the Precision--Recall (PR) 
		curve, enabling an objective assessment of the trade-off between false positives and missed 
		detections.

\begin{figure}[H]
			
			\hspace{-0.15cm}\includegraphics[width=0.58\textwidth]{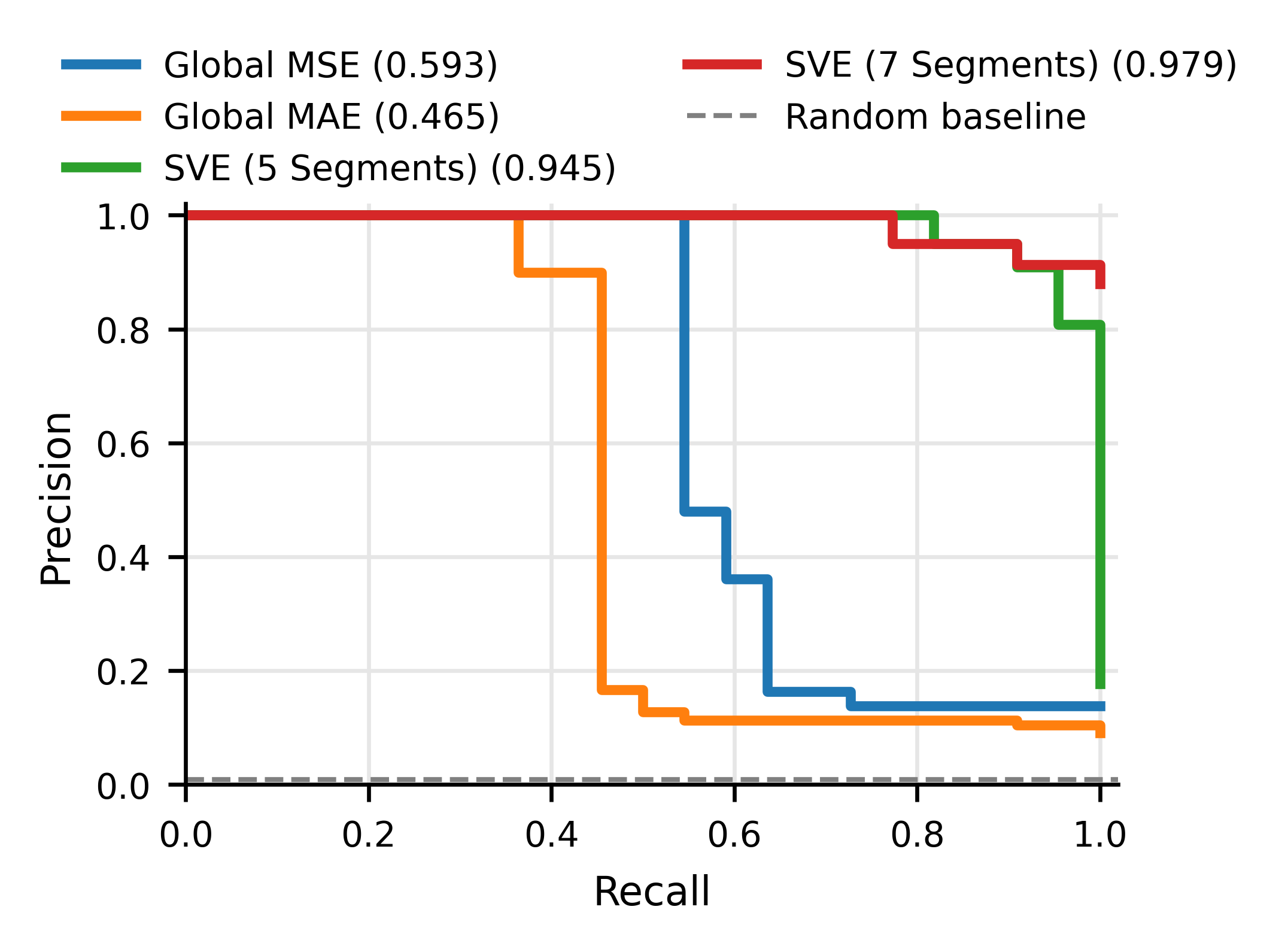}
			\caption{Precision--Recall curves for different reconstruction-based anomaly scores.}
			\label{fig:pr-auc}
		
		\end{figure}

		Figure~\ref{fig:pr-auc} illustrates the PR curves for different anomaly scoring strategies.
		In addition to global reconstruction metrics, two SVE configurations are reported. 
		{A repeated nested stratified five-fold cross-validation sensitivity analysis was conducted exclusively 
		on the validation set to assess the SVE configuration without using samples from the independent test set.
		The inner folds were used to evaluate the number of vertical segments and the number of aggregated segments, 
		whereas the outer folds were used to assess the robustness of the selected configuration.}
		The analysis identified segmentations with five and seven vertical 
		regions as the most promising configurations, both achieving nearly identical mean PR-AUC 
		values on the validation folds. Since the observed difference was negligible and within the 
		variation across folds, both configurations were subsequently evaluated on the independent test set.
		
		\vspace{-5pt}

		Table~\ref{tab:broach_detection_performance} confirms the advantage of localized anomaly 
		scoring. While global MSE and MAE achieve high precision, their substantially lower recall 
		indicates that localized anomaly patterns are diluted when reconstruction errors are aggregated 
		over the entire image.

\begin{table}[H]		
		\caption{Detection performance of global and SVE-based anomaly scoring methods on the broaching dataset test set.}			
			\label{tab:broach_detection_performance}
			\begin{tabularx}{\textwidth}{lCCCC}
\toprule
\textbf{Method} & \textbf{PR-AUC }& \textbf{F1-Score} & \textbf{Precision} & \textbf{Recall} \\
\midrule
Global MSE & 0.593 & 0.667 & 1.000 & 0.500 \\
Global MAE & 0.465 & 0.516 & 0.889 & 0.364 \\
SVE (5 Segments) & 0.945 & 0.909 & 0.909 & 0.909 \\
SVE (7 Segments) & 0.979 & 0.850 & 0.944 & 0.773 \\
\bottomrule
\end{tabularx}
			
		\end{table}

		The sensitivity analysis demonstrates that the choice of the segmentation configuration has a substantial 
		impact on anomaly detection performance. This observation suggests that defect signatures are concentrated 
		within localized spectrogram regions rather than being uniformly distributed across the complete image. 
		Consequently, evaluating reconstruction errors locally improves the detection of short-lived vibration anomalies. 
		The results further highlight the importance of performing a dedicated segment-sensitivity analysis, as the 
		effectiveness of localized anomaly scoring strongly depends on the chosen spatial partitioning. Overall, the 
		SVE-based approaches outperform the global reconstruction metrics, demonstrating that localized 
		reconstruction-error evaluation is an effective strategy for detecting vibration anomalies that appear as 
		vertically concentrated structures in spectrogram representations.

		\subsection{Comparison with Baseline Methods}

		The proposed method is compared against the classical baseline approaches introduced in Section \ref{baseline}.
		{All methods use representations preserving the information from the same three vibration axes, thereby avoiding 
		differences in available input information between the CNN-AE and the classical baselines.}

\begin{figure}[H]
			\includegraphics[width=0.58\textwidth]{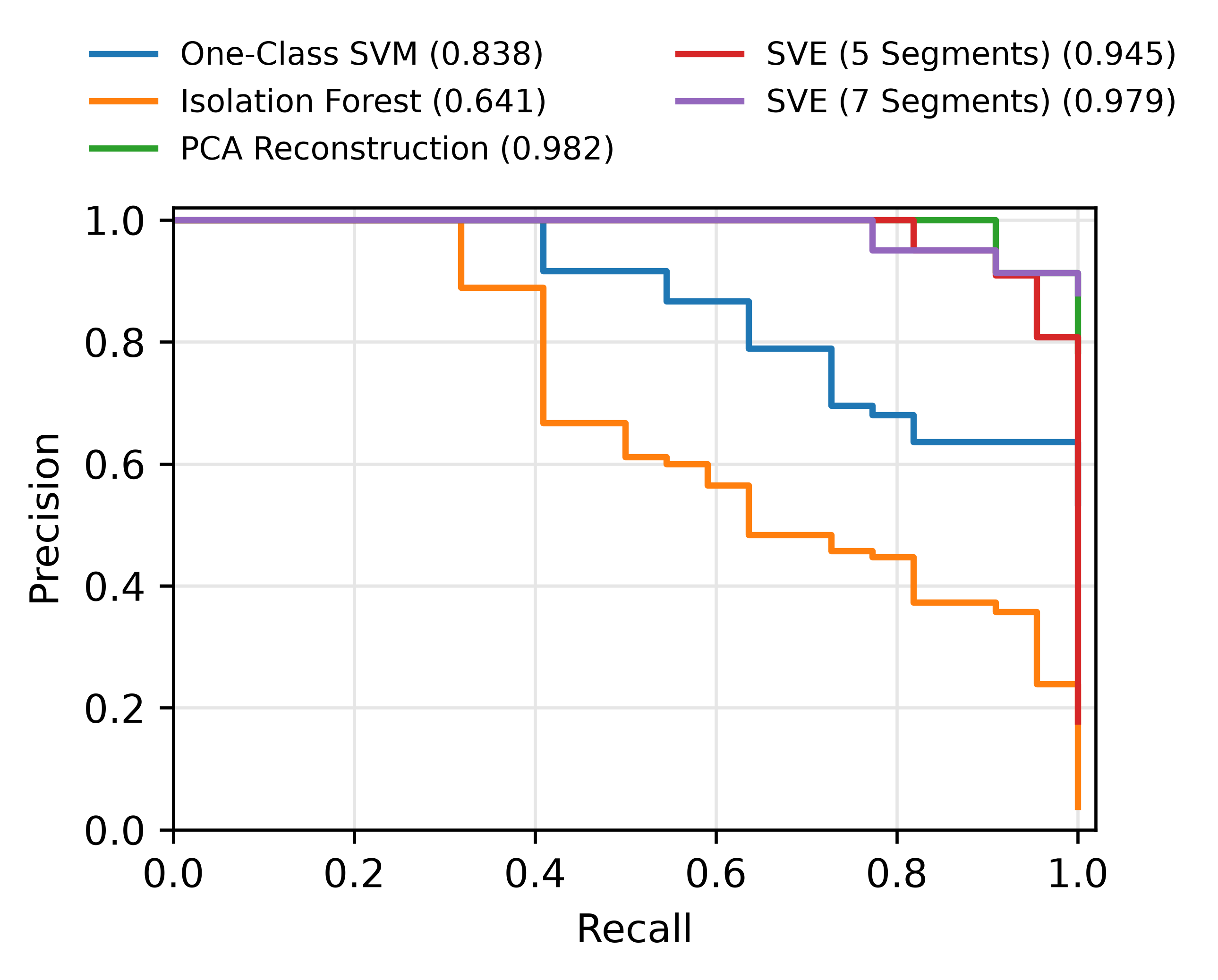}
			\caption{Precision--Recall curves for the proposed method and baseline approaches.}
			\label{fig:pr_curve_baselines}
		\end{figure}

		The results of Table~\ref{tab:broach_baseline_metrics} and Figure~\ref{fig:pr_curve_baselines} show that 
		the proposed SVE-based approach achieves {competitive performance relative to} the considered baseline methods.
		In particular, PCA reconstruction provides strong results and represents the most competitive 
		baseline on the broaching test set.

		\begin{table}[H]
			\caption{Comparison
 of detection methods on the broaching test set.}
			\label{tab:broach_baseline_metrics}
			\begin{tabularx}{\textwidth}{lRRRR}
\toprule
\textbf{Method} & \textbf{PR-AUC} & \textbf{F1-Score} & \textbf{Precision} & \textbf{Recall} \\
\midrule
Isolation Forest & 0.641 & 0.531 & 0.481 & 0.591 \\
One-Class SVM & 0.838 & 0.700 & 0.778 & 0.636 \\
PCA Reconstruction & 0.982 & 0.842 & 1.000 & 0.727 \\
Global MSE & 0.593 & 0.667 & 1.000 & 0.500 \\
Global MAE & 0.465 & 0.516 & 0.889 & 0.364 \\
SVE (5 Segments) & 0.945 & 0.909 & 0.909 & 0.909 \\
SVE (7 Segments) & 0.979 & 0.850 & 0.944 & 0.773 \\
\bottomrule
\end{tabularx}
		\end{table}

\vspace{-12pt}

		In contrast, the one-class SVM and isolation forest show lower detection performance, particularly 
		with respect to PR-AUC. Furthermore, both SVE configurations remain competitive with 
		the considered baseline methods, indicating that the proposed localized anomaly scoring strategy is 
		robust even when its parameters are selected without access to the test data.
	
		\subsection{Sensitivity of SVE to Segment Configuration}

		To assess the robustness of the proposed Segmented Vertical Error (SVE), a systematic sensitivity analysis 
		was conducted to investigate the influence of its configuration parameters on anomaly detection performance.

\begin{figure}[H]
			\includegraphics[width=0.58\textwidth]{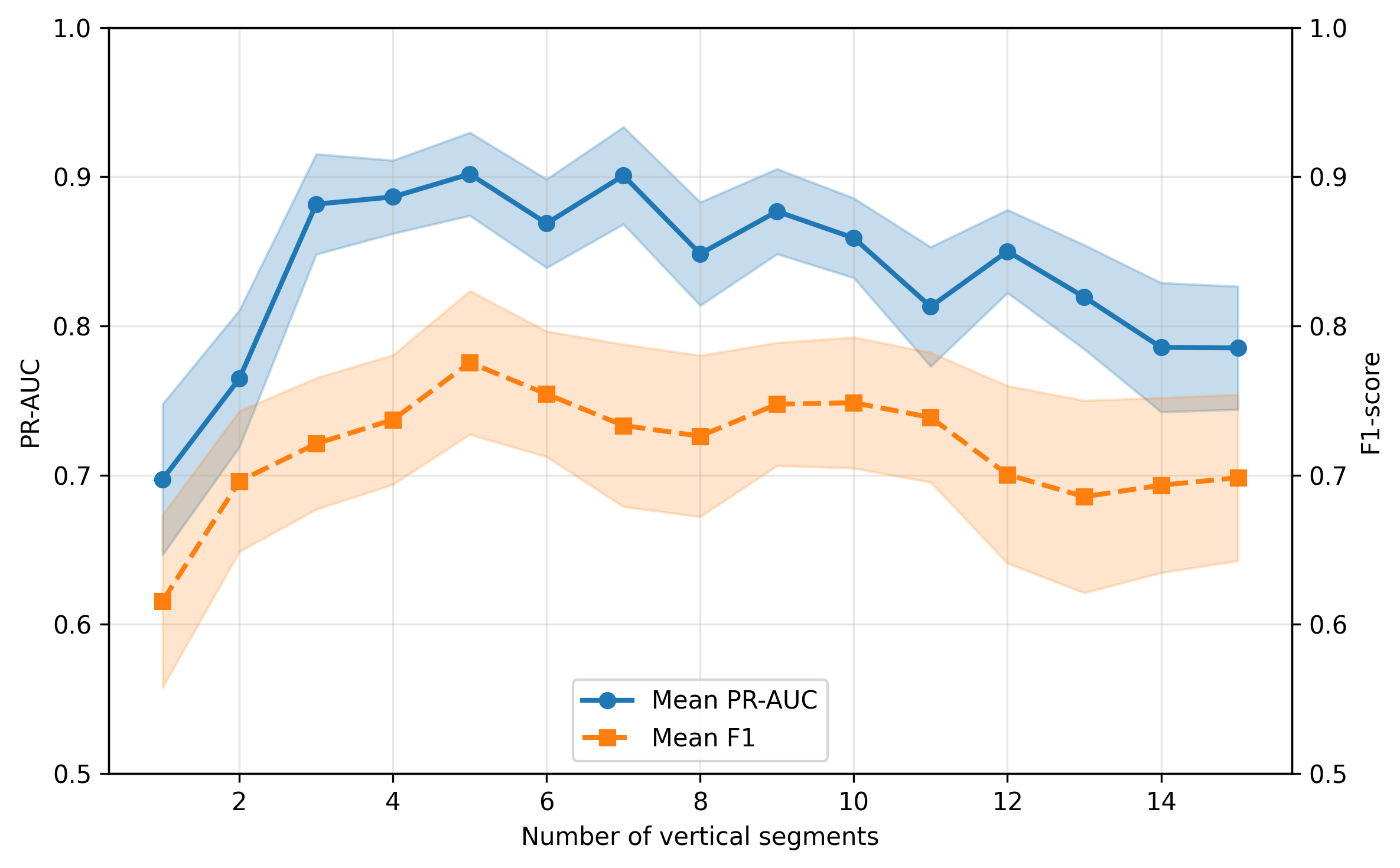}
			\caption{Sensitivity of SVE to the number of vertical segments measured by PR-AUC and F1-score. 
			The shaded regions indicate the standard deviation across cross-validation folds.}
			\label{fig:sve_sensitivity}
		
		\end{figure}

		To reduce the influence of a specific validation split, a cross-validation-based sensitivity analysis was 
		performed on the validation dataset. The number of vertical segments was varied between 1 and 15, while 
		different top-$k$ aggregation strategies were evaluated. 
		{The independent test set was not used during this analysis, and the selected SVE configuration was subsequently 
		applied unchanged in the primary fixed-split evaluation.}
		Figure~\ref{fig:sve_sensitivity} reports the mean PR-AUC and mean F1-score together with their standard 
		deviations across the cross-validation folds.

		The results indicate that the segmentation strategy influences detection performance, although no single 
		configuration consistently dominates across all cross-validation folds. Several neighboring segment 
		counts achieve comparable PR-AUC values, suggesting that the proposed SVE metric is reasonably robust 
		with respect to moderate changes in the segmentation.

		This finding implies that the main performance improvement stems from localized reconstruction error 
		evaluation compared to global image-level metrics, while the exact segment count represents a secondary 
		design parameter that should nevertheless be evaluated systematically.

		Furthermore, the evaluation showed that increasing the top-$k$ parameter did not improve 
		detection performance. The best results were consistently obtained for top-$k=1$, suggesting that 
		localized high-error regions contain the most relevant anomaly information, whereas averaging 
		reconstruction errors across multiple segments tends to dilute the anomaly signal.

		\subsection{Evaluation on a Public Turning Dataset} 
		\label{validation}
		
		To evaluate the {cross-process applicability} of the proposed anomaly detection pipeline to another machining 
		process, an additional validation experiment was conducted using the publicly available 
		Turning Dataset for Chatter Diagnosis Using Machine Learning
 		by Khasawneh et al., hosted by Mendeley Data \cite{KhasawnehDataset}.
		
		The dataset contains measurements acquired during turning experiments under varying spindle 
		speeds and tool overhang configurations. These operating conditions were intentionally selected 
		to induce unstable cutting behavior and chatter vibrations. Although multiple sensor modalities 
		are available, only the vibration signal recorded at the tool holder was used in this study, as 
		it directly reflects dynamic process instabilities during machining.

		The resulting dataset provides an external proof-of-concept evaluation of whether the proposed 
		spectrogram-based workflow can be applied to a machining process beyond the original broaching 
		application.
				
		\subsubsection{Data Preparation}

			The original turning dataset consists of vibration recordings with varying durations. 
			While the dataset provides experiment-level information regarding stable and chatter-prone 
			cutting conditions, the proposed anomaly detection workflow requires fixed-size input samples. 
			Therefore, each recording was segmented into non-overlapping windows of 2.5 s. 

			The window duration of 2.5 s was chosen as a practical compromise between the information 
			content of individual samples and the number of samples available for autoencoder training. 
			Shorter windows were considered too limited to provide sufficiently informative spectrogram 
			representations, whereas longer windows would have substantially reduced the number of available 
			training samples. The selected duration therefore provided a practical balance between sample 
			informativeness and dataset size.

			This segmentation increased the number of available samples and enabled the isolation of stable 
			and unstable process phases that may coexist within a single recording. 
			Prior to segmenting the recordings, the original experiment-level labels were used to derive 
			the window-labeling strategy. 

			FFT analyses of complete recordings were performed to identify spectral characteristics 
			associated with chatter and stable cutting conditions. Based on these observations, 
			thresholds for peak detection and harmonic-pattern identification were defined. 
			These criteria were subsequently applied to each 2.5-s window individually. 
			This procedure enabled the generation of localized labels at window level, allowing stable 
			and unstable cutting phases to be separated even when both occurred within the same recording.
		
			For each window, a fast Fourier transform (FFT) was computed, and the resulting spectrum 
			was examined for harmonic peak structures characteristic of regenerative chatter vibrations. 
			Windows exhibiting a pronounced harmonic pattern within the frequency range below 5~kHz were 
			assigned the label ``chatter'',
			whereas all remaining windows were labeled ``no chatter''. 
			{Because these window-level labels are derived from frequency-domain characteristics of the same 
			vibration signals that are subsequently represented as STFT spectrograms, the labeling and 
			model-input representations are not fully independent. The turning experiment should therefore 
			be interpreted as a cross-process proof of concept using frequency-derived window labels rather 
			than as an independently labeled external benchmark.}
			The labeled vibration windows were subsequently transformed into spectrograms using the 
			preprocessing pipeline described in Section~\ref{signalProcessing}. 
			{The resulting spectrogram images were evaluated using run-wise grouped five-fold cross-validation, 
			with the machining run used as the grouping variable. All windows originating from the same machining 
			run were assigned to the same outer fold, thereby preventing run-specific information from being 
			shared between model development and performance evaluation. In each cross-validation iteration, 
			one outer fold was held out for performance evaluation, while the remaining four folds formed the 
			development set. Within this development set, the machining runs were further separated into 
			model-training and threshold-calibration subsets using a group-wise split. The autoencoder and 
			the classical one-class methods were trained exclusively on nominal samples from the respective 
			training runs. The labeled calibration runs were used to determine the decision threshold, 
			which was subsequently applied unchanged to the held-out outer fold. This procedure was repeated 
			until each outer fold had served once as the held-out evaluation fold. The same outer-fold assignments 
			and calibration protocol were used for the CNN autoencoder and all classical baseline methods. 
			This run-wise evaluation differs from the fixed-split protocol used for the industrial broaching 
			dataset because several turning windows originate from the same machining run and are therefore 
			not statistically independent at the run level.}

			For reproducibility, the executable preprocessing implementation used for the public turning dataset
			sets the sampling frequency to 50~kHz and processes the 2.5-s windows without additional padding.
			It calls \texttt{scipy.signal.spectrogram} in PSD mode with 512-sample segments and 384-sample overlap,
			corresponding to a 128-sample hop. The generator does not explicitly pass a window function; therefore,
			SciPy's default Tukey window with $\alpha=0.25$ is used. The other effective defaults are constant detrending,
			a one-sided spectrum, density scaling, and $n_{\mathrm{fft}}=512$ (the \texttt{nperseg} value).
			Only frequencies up to 5~kHz are retained. The PSD is converted using
			$10\log_{10}(S_{xx}+10^{-12})$, clipped to $[-75,-20]$~dB, and linearly normalized with an inverted
			range. The three axis spectrograms are stacked as RGB channels (X/Y/Z), vertically flipped, and resized
			to 150 $\times$ 100 pixels using the image-library default nearest-neighbor interpolation.

			These settings are summarized in Table~\ref{tab:turning_stft_params}.

		\begin{table}[H] 
			 
			\caption{STFT and image-generation parameters for the public turning dataset.} 
			\label{tab:turning_stft_params}
			 \begin{tabularx}{\textwidth}{LL} \toprule \textbf{Parameter} & \textbf{Value} \\ \midrule Sampling rate & 
				\(50\,\mathrm{kHz}\) \\ Input window & \(2.5\,\mathrm{s}\) \\ Window function & Tukey (\(\alpha = 0.25\)) 
				\\ Segment length & \(512\) samples \\ FFT length & \(512\) samples \\ Overlap/hop length 
				& \(384 / 128\) samples (\(75\,\%\) overlap) \\ Detrending & Constant \\ Spectrum & One-sided PSD, density scaling 
				\\ Retained frequency range & \(0\text{--}5\,\mathrm{kHz}\) \\ PSD-to-dB offset & \(10^{-12}\) 
				\\ Clipping range & \([-75,-20]\,\mathrm{dB}\) \\ Image conversion & Inverted linear normalization, vertical flip 
				\\ Channels & RGB (X/Y/Z) \\ Output resolution & \(150 \times 100\) pixels (width \(\times\) height) 
				\\ \bottomrule 
			\end{tabularx} 
		\end{table}

\begin{figure}[H]
			\includegraphics[width=0.53\textwidth]{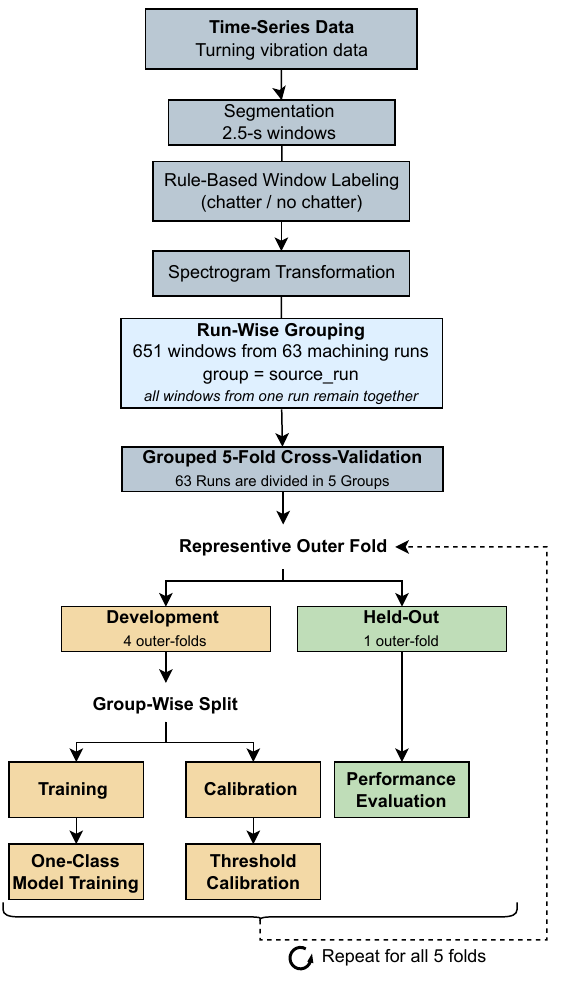}
			\caption{Processing and run-wise grouped five-fold evaluation protocol for the turning dataset. 
			Windows originating from the same machining run remain grouped throughout model training, 
			threshold calibration, and performance evaluation.}
			\label{fig:dataset_pipeline}
		\end{figure}

		Figure~\ref{fig:dataset_pipeline} summarizes the preprocessing and run-wise cross-validation 
		procedure applied to the turning dataset.		
		
		\subsubsection{Detection Performance on the Turning Dataset}
		
\begin{figure}[H]
			
			\renewcommand{\arraystretch}{1.3}
			
			\begin{tabular}{c c}
				
				& \textbf{\hspace{0.75cm}Input \hspace{1.5cm} Reconstruction \hspace{1.5cm} Residual} \\[0.0cm]
				
				\multirow{-7}{*}{\textbf{Normal}} &
				\includegraphics[width=0.75\textwidth]{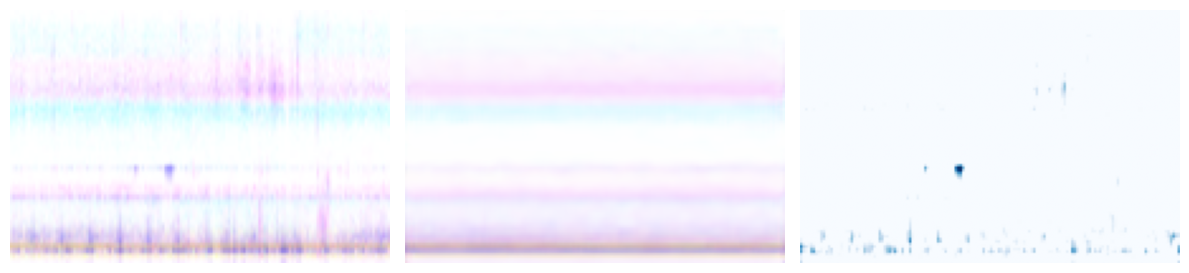} \\[0.4cm]
				
				\multirow{-7}{*}{\textbf{Anomaly}} &
				\includegraphics[width=0.75\textwidth]{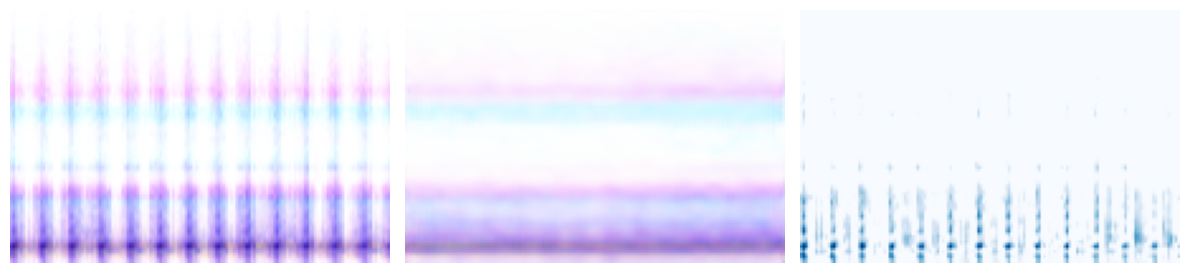}
				
			\end{tabular}
			\caption{Qualitative
 reconstruction results on the turning dataset. 
			Anomalous samples exhibit clearly visible reconstruction errors in the residual images.}
			\label{fig:validation_reconstruction}
			
		\end{figure}

		{Anomaly detection was performed using the reconstruction-error-based scoring mechanism described 
		previously. For each outer fold, the development runs were separated into model-training and 
		threshold-calibration subsets using a group-wise split. The CNN-AE was trained exclusively on 
		nominal samples from the respective training runs, while the decision thresholds were calibrated 
		using the labeled calibration runs. The resulting thresholds were subsequently applied unchanged 
		to the held-out outer fold. Performance was summarized as the mean and standard deviation across 
		the five outer folds, as reported in Table~\ref{tab:turning_method_comparison}.} {Because of the limited size of this public dataset and the use of 
		frequency-derived window labels, the resulting scores should be interpreted as an external 
		cross-process proof of concept rather than as a fully independent benchmark.}
		Representative reconstruction results on the turning dataset are shown in Figure~\ref{fig:validation_reconstruction}.
		
		\vspace{-6pt}

		\begin{table}[H]
		\small
		\caption{Anomaly-detection
 performance on the turning dataset using run-wise grouped five-fold cross-validation (mean $\pm$ standard deviation).}
		\label{tab:turning_method_comparison}
		\begin{tabularx}{\textwidth}{Lrrrrrrrr}
\toprule
\textbf{\multirow{2.5}{*}{Method}} & \multicolumn{2}{c}{\textbf{PR-AUC}} & \multicolumn{2}{c}{\textbf{F1}} & \multicolumn{2}{c}{\textbf{Precision}} & \multicolumn{2}{c}{\textbf{Recall}} \\ 
\cmidrule{2-9}
 & \textbf{Mean} & \textbf{SD} & \textbf{Mean} & \textbf{SD} & \textbf{Mean} & \textbf{SD} & \textbf{Mean} & \textbf{SD} \\
\midrule
CNN-AE Global MAE & 0.819 & 0.230 & 0.634 & 0.267 & 0.699 & 0.140 & 0.657 & 0.387 \\
CNN-AE Global MSE & 0.788 & 0.221 & 0.603 & 0.269 & 0.633 & 0.202 & 0.657 & 0.387 \\
CNN-AE SVE Max & 0.775 & 0.215 & 0.633 & 0.224 & 0.705 & 0.222 & 0.731 & 0.354 \\
CNN-AE SVE Top-k & 0.788 & 0.219 & 0.581 & 0.263 & 0.592 & 0.223 & 0.657 & 0.387 \\
Isolation Forest & 0.882 & 0.138 & 0.747 & 0.193 & 0.879 & 0.132 & 0.716 & 0.279 \\
One-Class SVM & 0.754 & 0.172 & 0.647 & 0.169 & 0.702 & 0.231 & 0.742 & 0.315 \\
PCA Reconstruction & 0.724 & 0.192 & 0.614 & 0.165 & 0.661 & 0.234 & 0.709 & 0.308 \\
\bottomrule
\end{tabularx}
		
		\end{table}

	{The run-wise grouped evaluation on the public turning dataset provides preliminary evidence of cross-process 
	applicability while also revealing substantial fold-to-fold variability. Isolation forest achieved the 
	strongest mean performance among the evaluated methods, whereas the CNN-AE with global MAE remained 
	competitive with one-class SVM and PCA reconstruction. The localized SVE scores did not consistently 
	outperform the global reconstruction-error metrics on the turning dataset. These findings indicate that 
	the benefit of localized reconstruction-error aggregation depends on the anomaly structure and should 
	not be interpreted as a universal advantage across machining processes.}

	{Among the classical baselines, isolation forest achieved the strongest overall performance, with a 
	mean PR-AUC of $0.882 \pm 0.138$ and a mean F1-score of \mbox{$0.747 \pm 0.193$}. One-class SVM achieved a 
	mean PR-AUC of $0.754 \pm 0.172$ and an F1-score of \mbox{$0.647 \pm 0.169$}, whereas PCA reconstruction 
	reached $0.724 \pm 0.192$ and $0.614 \pm 0.165$, respectively. Among the CNN-AE scoring strategies, 
	global MAE achieved the highest mean PR-AUC (\mbox{$0.819 \pm 0.230$}), while the F1-scores of global MAE 
	($0.634 \pm 0.267$) and SVE MAX ($0.633 \pm 0.224$) were comparable. In contrast to the industrial 
	broaching results, the localized SVE scores therefore did not provide a consistent advantage over 
	global reconstruction errors on the turning dataset. The comparatively large standard deviations 
	across several metrics further indicate substantial variation between the held-out machining runs.}

	{The industrial broaching dataset presents a different evaluation scenario, with more heterogeneous 
	anomaly appearances and an independent fixed test set. In this setting, the localized SVE scores 
	substantially improved detection performance relative to the global CNN-AE reconstruction-error metrics. 
	However, classical methods, particularly PCA reconstruction, remained competitive. This indicates that 
	the main benefit demonstrated by SVE lies in the localized evaluation of the autoencoder reconstruction 
	error rather than in a general superiority of the CNN-AE over classical anomaly-detection methods.}

	{\textls[-10]{Taken together, these results provide preliminary evidence that the spectrogram-based processing and 
	anomaly-detection workflow can be applied across different machining processes. The turning experiment 
	demonstrates cross-process applicability under a run-wise grouped evaluation, while also showing that 
	simple classical methods can remain highly competitive depending on the anomaly structure. 
	The principal contribution of the proposed SVE strategy is therefore its improvement over global 
	reconstruction-error evaluation for localized anomaly patterns, as demonstrated on the industrial 
	broaching dataset, rather than a universal performance advantage over classical anomaly-detection methods.}}				
	
	\section{Discussion}
	
	The reconstruction results demonstrate that the proposed autoencoder successfully captures the 
	dominant time--frequency characteristics of normal process behavior, while vibration anomalies 
	produce structured residual patterns. This supports the suitability of reconstruction-based 
	anomaly detection for vibration monitoring in machining processes.
	
	The obtained PR-AUC values indicate a strong separation between normal and anomalous samples despite 
	the severe class imbalance present in the industrial dataset.
	
	A key extension of the proposed approach is the introduction of the Segmented Vertical Error (SVE). 
	By evaluating reconstruction errors on localized vertical segments, the method explicitly 
	incorporates prior knowledge about the structure of vibration-induced anomalies. 
	This allows for a more robust detection of relevant patterns while reducing the influence of 
	horizontally distributed artifacts. 
	In addition, the segment-wise evaluation enables a coarse localization of anomalies within the 
	spectrogram, which may provide further insights into process dynamics.

	The sensitivity analysis showed that SVE remains effective across a broad range of segment configurations. 
	No single segmentation setting consistently dominated across all cross-validation folds, and several 
	neighboring configurations achieved comparable PR-AUC values. This suggests that the observed performance 
	improvement is primarily attributable to the localized reconstruction-error evaluation principle itself 
	rather than to a highly tuned segment configuration. The results further indicate that a systematic 
	sensitivity analysis is beneficial for identifying suitable segmentations while confirming the robustness 
	of the proposed SVE formulation.
	
	{The run-wise grouped evaluation on the public turning dataset provides preliminary evidence of cross-process 
	applicability, while the substantial fold-to-fold variability and competitive performance of classical methods indicate 
	that the benefit of localized reconstruction-error aggregation depends on the anomaly structure. Moreover, because the 
	window-level labels were derived from frequency-domain characteristics of the same vibration signals used for the spectrogram 
	inputs, the turning experiment should be interpreted as an external cross-process proof of concept rather than as an 
	independently labeled validation of generalization.}
	
	{From an industrial edge-deployment perspective, the computational feasibility of the proposed framework warrants 
	explicit consideration. The benchmark results show that the dominant computational cost arises from signal processing, 
	including the multi-channel Short-Time Fourier Transform (STFT) and RGB spectrogram generation. Across the evaluated 
	hardware platforms, this preprocessing stage accounts for the majority of the overall pipeline latency, whereas CNN-AE 
	inference and subsequent anomaly scoring represent only a small fraction of the total processing time.
 	Consequently, the computational requirements of the proposed pipeline are primarily determined by spectrogram generation 
	rather than neural-network inference. This finding indicates that optimization efforts for real-time deployment should 
	primarily target the signal-processing stage. However, because the inference 
	latency of the classical baseline models was not measured separately, these results should not be interpreted as a 
	direct runtime comparison between the CNN-AE and the baseline methods.}

	{From a maintenance perspective, the proposed method should be interpreted as an anomaly-detection component 
	rather than as a complete maintenance-decision system. An elevated anomaly score may trigger further inspection or 
	diagnostic activities, but the method does not identify root causes, estimate component degradation, or prescribe 
	maintenance actions. Such decisions require additional diagnostic and process information and may be incorporated into 
	higher-level inspection and maintenance strategies \cite{Wei2026-CM}.}

	Nevertheless, several limitations must be considered. 
	First, the approach relies on the assumption that the training data accurately represent normal 
	process behavior. 
	Any contamination with anomalous samples may negatively affect the learned representation. 
	Second, the selection of the decision threshold remains application-dependent and requires careful 
	tuning based on the desired trade-off between false alarms and missed detections. \
	{Third, the retrospective annotation of the industrial broaching dataset introduces a potential source of observer dependence. 
	The characteristic vibration patterns used for annotation were established from representative comparisons of 
	synchronously acquired force and acceleration measurements. The individual evaluation samples were subsequently 
	annotated by a single expert based on the corresponding vibration spectrograms. Since not every annotation was 
	additionally cross-checked against the corresponding force signal, some degree of observer dependence cannot be excluded. 
	Future studies could further strengthen the reference annotations through additional process information or independent \mbox{expert assessments.}}

	Furthermore, although the proposed SVE metric was found to be reasonably robust across a range of 
	segment configurations, the {applicability of the selected segmentation} to other processes and machine 
	tools remains an open question and should therefore be assessed through a dedicated sensitivity analysis. 
	Finally, the method operates on precomputed spectrograms and does not explicitly consider temporal 
	dependencies across multiple process cycles. 
	
	Despite these limitations, the proposed framework provides a practical basis for one-class 
	anomaly detection in machining processes. The autoencoder can be trained without labeled fault data, 
	although threshold calibration in the present study uses labeled validation data. The evaluation on 
	industrial broaching data and the public turning dataset highlights the potential of the approach 
	for industrial condition-monitoring applications.
	
	\section{Conclusions} 
	This work presented and evaluated a spectrogram-based {one-class} anomaly detection framework for machining vibration 
	signals.
	A particular contribution of this study is its evaluation on an industrial broaching dataset with rare process anomalies.
	The anonymized broaching spectrograms and associated expert annotations made available with this study provide the engineering
	community with an industrial test case for developing and comparing vibration-based anomaly detection methods under severe class imbalance. 
	
	By transforming multivariate vibration signals into spectrogram representations, the approach enables 
	convolutional autoencoders to learn characteristic patterns of normal process behavior. 
	The results indicate that reconstruction-based anomaly detection can identify chatter-related process instabilities under highly imbalanced conditions. 
	The proposed Segmented Vertical Error (SVE) improves anomaly localization by emphasizing localized 
	reconstruction deviations that are characteristic of vibration-induced anomalies. Evaluation on the industrial broaching data showed a 
	clear performance advantage of localized SVE scoring over global CNN-AE reconstruction-error metrics, 
	{while classical approaches, particularly PCA reconstruction, remained competitive.}
	Furthermore, the cross-validation-based sensitivity analysis demonstrated that localized reconstruction-error 
	evaluation remained effective across a range of segment configurations, indicating that the observed performance 
	gain on the broaching dataset was not restricted to a single highly tuned parameter setting.
	In addition, the application to a public turning 
	dataset provides preliminary evidence that the proposed spectrogram-based workflow may extend beyond 
	the original broaching scenario, although broader claims of generalization require further cross-machine 
	and cross-process validation. Future work may focus on integrating temporal information across 
	multiple process cycles, investigating adaptive SVE configurations, and extending the approach 
	towards real-time deployment in industrial environments.

\vspace{6pt}			
\section*{Author Contributions}
Conceptualization, S.Z. and C.G.; methodology, S.Z.; software, S.Z.; validation, S.Z. and C.G.; formal analysis, S.Z.; investigation, S.Z.; resources, S.Z. and C.G.; data curation, S.Z.; writing--original draft preparation, S.Z.; writing--review and editing, S.Z. and C.G.; visualization, S.Z.; supervision, C.G.; project administration, S.Z.; funding acquisition, C.G. All authors have read and agreed to the published version of the manuscript.

\section*{Funding}
This article is funded by the Open Access Publication Fund of South Westphalia University of Applied Sciences.

\section*{Data Availability}
The source code, experimental pipeline, trained models, and all data that can be publicly disclosed are available in the companion repository: \url{https://github.com/fhswf/spectrogram-anomaly-ae}. Certain industrial raw datasets used in this study cannot be released without restriction due to industrial confidentiality and proprietary agreements. Publicly shareable datasets, processing scripts, experiment configurations, and documentation required to reproduce the presented results are provided through the companion repository.

\section*{Acknowledgments}
During the preparation of this manuscript, the authors used OpenAI Codex 26.51007.21434 with GPT-5.5 for the purposes of obtaining editorial feedback, identifying unclear passages, and suggesting improvements in organization and wording. The authors have reviewed and edited the output and take full responsibility for the content of this publication. The authors would like to thank Forst Technologie GmbH for supporting this research and providing the industrial environment in which significant parts of this work were conducted. We further express our sincere gratitude to Patrik Vogtel for his valuable guidance, constructive feedback, and technical expertise.

\section*{Conflicts of Interest}
Sebastian Zeleny is employed by Forst Technologie GmbH \& Co. KG, which provided the industrial dataset used in this study. The remaining author declares that the research was conducted in the absence of any commercial or financial relationships that could be construed as a potential conflict of interest.

\appendix
\section{Complete CNN Autoencoder Specification}
\label{app:ae_architecture}

\textls[-10]{The following tables document the reference CNN autoencoder used in the fixed-split
experiments. Input and output dimensions are reported as height $\times$ width
$\times$ channels, excluding the batch dimension. All convolutional layers use a
single bias per output channel. The parameter counts are those reported by Keras and
sum to 251{,}383 trainable parameters.}

\begin{table}[H] 
	 \small
	\caption{Layer-wise
 specification of the reference convolutional autoencoder with a bottleneck dimension of 16. A dash indicates that the corresponding property is not applicable.} 
	\label{tab:autoencoder_complete} 
	
\begin{adjustwidth}{-\extralength}{0cm}
\begin{tabularx}{\fulllength}{llcCCClr} 

		\toprule \textbf{Stage} & \textbf{Layer} & \textbf{Kernel/Size} & \textbf{Stride} & \textbf{Padding} & \textbf{Activation} & \textbf{Output Shape} & \textbf{Parameters} \\ 
		\midrule Input & Input & -- & -- & -- & -- & \(100 \times 150 \times 3\) & 0 \\ 
		\addlinespace Encoder & Conv2D (4) & \(3 \times 3\) & \(1 \times 1\) & same & ReLU & \(100 \times 150 \times 4\) & 112 \\ 
		& MaxPooling2D & \(2 \times 2\) & \(2 \times 2\) & same & -- & \(50 \times 75 \times 4\) & 0 \\ & Conv2D (8) & \(3 \times 3\) 
		& \(1 \times 1\) & same & ReLU & \(50 \times 75 \times 8\) & 296 \\ & MaxPooling2D & \(2 \times 3\) & \(2 \times 3\) 
		& same & -- & \(25 \times 25 \times 8\) & 0 \\ & Conv2D (12) & \(3 \times 3\) & \(1 \times 1\) & same & ReLU 
		& \(25 \times 25 \times 12\) & 876 \\ & Flatten & -- & -- & -- & -- & \(7500\) & 0 \\ 
		\addlinespace Bottleneck & Dense (16) & -- & -- & -- & ReLU & \(16\) & 120{,}016 \\ \addlinespace Decoder 
		& Dense (7500) & -- & -- & -- & ReLU & \(7500\) & 127{,}500 \\ & Reshape & \(25 \times 25 \times 12\) & -- & -- & -- & 
		\(25 \times 25 \times 12\) & 0 \\ & Conv2D (12) & \(3 \times 3\) & \(1 \times 1\) & same & ReLU & \(25 \times 25 \times 12\) 
		& 1308 \\ & UpSampling2D & \(2 \times 3\) & -- & -- & nearest & \(50 \times 75 \times 12\) & 0 \\ & Conv2D (8) & \(3 \times 3\) 
		& \(1 \times 1\) & same & ReLU & \(50 \times 75 \times 8\) & 872 \\ & UpSampling2D & \(2 \times 2\) & -- & -- & nearest 
		& \(100 \times 150 \times 8\) & 0 \\ & Conv2D (4) & \(3 \times 3\) & \(1 \times 1\) & same & ReLU & \(100 \times 150 \times 4\) 
		& 292 \\ & Conv2D (3) & \(3 \times 3\) & \(1 \times 1\) & same & sigmoid & \(100 \times 150 \times 3\) & 111 \\ 
		\midrule \multicolumn{7}{r}{Total 
} & 251{,}383 \\ 
		\bottomrule 
	\end{tabularx} 
	\end{adjustwidth}
\end{table}

\vspace{-10pt}
\begin{table}[H] 
	 
	\small 
	\setlength{\tabcolsep}{4pt} 
	\caption{Training and reproducibility settings of the convolutional autoencoder.} 
	\label{tab:autoencoder_training} 
	\begin{tabularx}{\columnwidth}{@{}lX@{}} \toprule \textbf{Item} & \textbf{Setting} \\ \midrule Training objective 
		& Self-reconstruction of nominal samples using MSE loss \\ Input preprocessing 
		& RGB images resized to \(150 \times 100\) pixels (width \(\times\) height), represented internally as \(100 \times 150 \times 3\), 
		and scaled to \([0,1]\) \\ Data split & \(80\,\%\) nominal training and \(20\,\%\) nominal validation; 
		shuffled with \texttt{random\_state=42} \\ Batch size & \(32\) \\ Optimizer & Adam with learning rate 
		\(1 \times 10^{-4}\), \(\beta_1 = 0.9\), \(\beta_2 = 0.999\), and \(\epsilon = 10^{-7}\); no weight decay; 
		AMSGrad disabled \\ Epochs (fixed split) & Maximum of \(4000\) epochs; the reference model was trained for 
		all \(4000\) epochs without early stopping \\ Early stopping (nested grouped CV) & Validation loss monitored with a 
		patience of \(250\) epochs and minimum improvement of \(10^{-6}\); best weights restored \\ Random seeds 
		& \(42\) for \texttt{PYTHONHASHSEED}, Python \texttt{random}, NumPy, TensorFlow, the nominal train/validation 
		split, and grouped-CV fold construction 
		\\ Software & Python 3.13.11; TensorFlow 2.21.0; Keras 3.14.1; NumPy 2.4.6; scikit-learn 1.9.0 
		\\ Hardware & Native Windows environment; no GPU detected; exact CPU model and RAM not recorded 
		\\ \bottomrule 
	\end{tabularx} 
\end{table}

The fixed-split training record and the nested grouped-CV implementation use the
same layer architecture and optimizer settings. They differ only in the stopping
protocol documented in Table~\ref{tab:autoencoder_training}: the fixed-split
reference model uses the full 4000-epoch budget, whereas grouped-CV fits use
validation-loss early stopping.

\section{Threshold Analysis of the SVE Method}
\label{app:sve_threshold}

\begin{figure}[H]
\begin{minipage}[t]{0.65\textwidth}
    \centering
    \includegraphics[width=\textwidth]{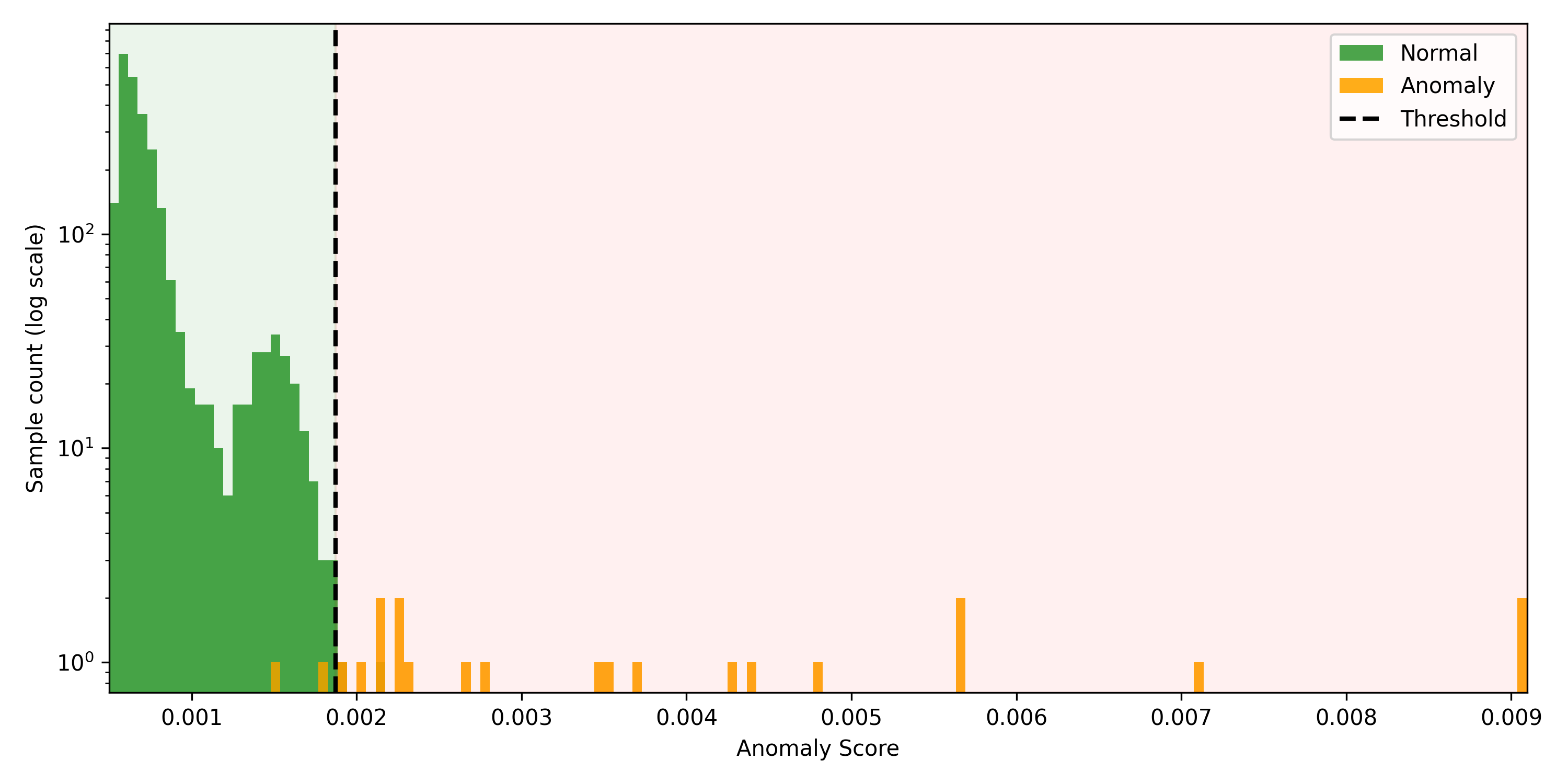}
\end{minipage}
\hfill
\begin{minipage}[t]{0.33\textwidth}
    \centering
    \includegraphics[width=\textwidth]{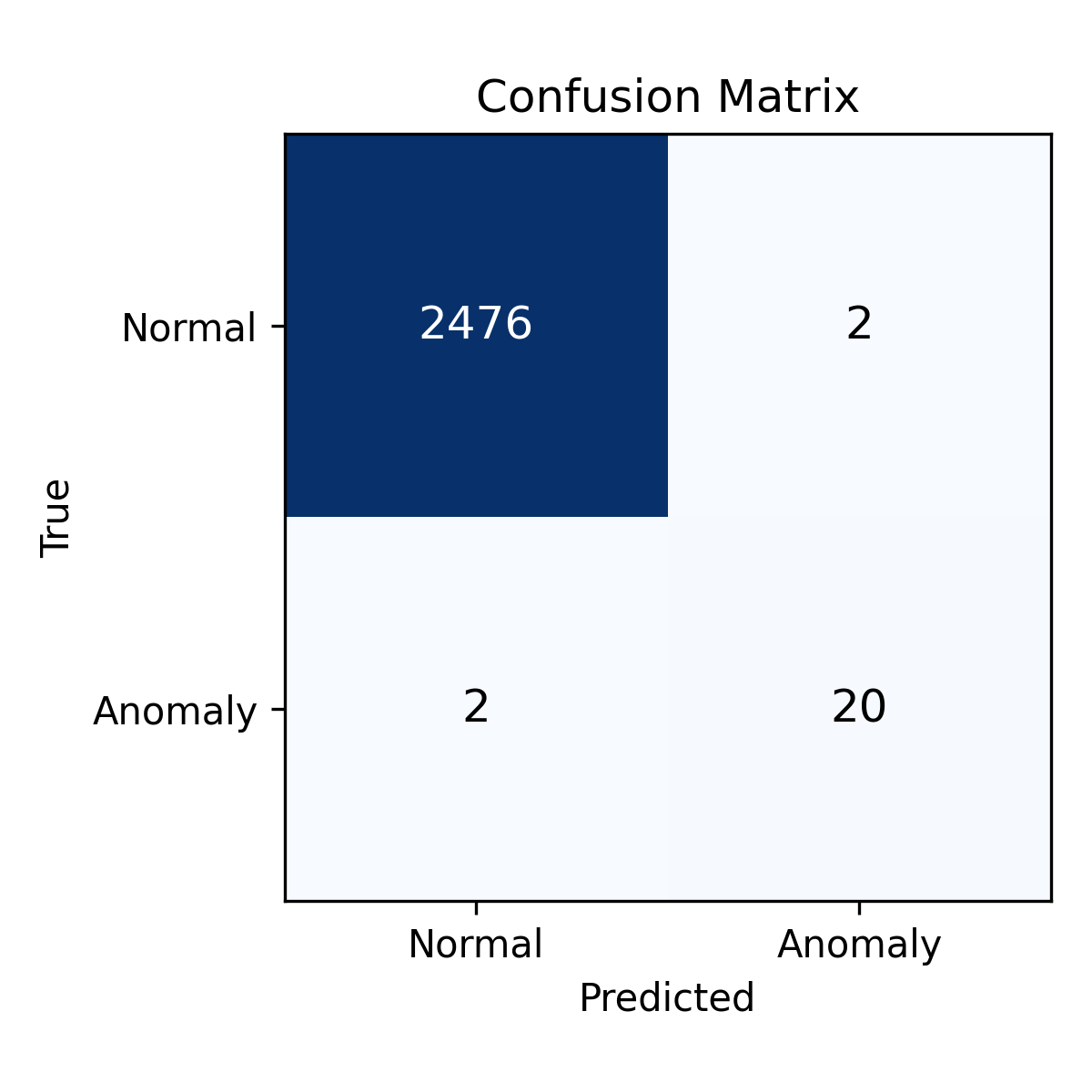}
\end{minipage}

\caption{
Distribution of SVE anomaly scores and resulting classification
performance. (\textbf{Left}) Histogram of anomaly scores for normal and anomalous
samples.
(\textbf{Right}) Corresponding confusion matrix obtained on the test set using
the selected threshold.
}
\label{fig:sve_threshold_analysis}
\end{figure}

Figure~\ref{fig:sve_threshold_analysis} illustrates the distribution of anomaly scores 
obtained with the SVE configuration consisting of five segments. The histogram 
provides an additional view of the score separation between normal and anomalous samples, 
while the confusion matrix summarizes the classification performance achieved using the 
selected decision threshold.

\vspace{-4pt}

\section{Cross-Validation and Robustness Analysis of Broaching Models}
\label{app:broaching_cv}

To assess the sensitivity of the classical baseline operating points to the
choice of evaluation samples, we performed a secondary nested threshold
cross-validation analysis on the complete labeled broaching evaluation pool.
The pool contained 4955 normal and 45~anomalous samples. Five stratified
outer folds were constructed, each containing 1000~samples and nine
anomalies. For each outer fold, thresholds were selected by maximizing F1 on
the remaining 4000 samples, corresponding to four inner folds, and were
then applied to the held-out outer fold.

The baseline models were always fitted on the same 5000 nominal training
images. Thus, this experiment audits threshold selection rather than tuning
baseline hyperparameters or retraining on the evaluation pool. Since the
released broaching data do not provide machining-run or stroke identifiers,
the folds are stratified at sample level and the reported variability should
not be interpreted as a grouped estimate of run-level generalization.

\begin{table}[H]
\caption{Nested
 threshold five-fold cross-validation performance of classical baselines on the broaching evaluation pool. 
Values are mean $\pm$ standard deviation across the five outer folds.}
\label{tab:broach_baseline_nested_cv}
\begin{tabular}{lrrrr}
\toprule
\textbf{Method} & \textbf{Precision} & \textbf{Recall} & \textbf{F1-Score} & \textbf{PR-AUC} \\
\midrule
One-class SVM & 0.728 $\pm$ 0.099 & 0.822 $\pm$ 0.127 & 0.770 $\pm$ 0.098 & 0.881 $\pm$ 0.073 \\
Isolation Forest & 0.713 $\pm$ 0.205 & 0.556 $\pm$ 0.176 & 0.586 $\pm$ 0.089 & 0.730 $\pm$ 0.097 \\
PCA Reconstruction & 0.900 $\pm$ 0.141 & 0.822 $\pm$ 0.099 & 0.852 $\pm$ 0.089 & 0.915 $\pm$ 0.086 \\
\bottomrule
\end{tabular}

\end{table}

The autoencoder robustness analysis differs from a classical model-level cross-validation. 
The CNN autoencoder was trained only once on the nominal training set and its reconstruction scores were kept fixed. 
The repeated cross-validation procedure was applied exclusively to the evaluation pool in order to assess the stability 
of threshold selection and performance metrics under different validation and test set compositions. Consequently, the 
results quantify score robustness rather than variability induced by model retraining.

\begin{table}[H]
\caption{Sensitivity
 of autoencoder anomaly score performance to different evaluation-set compositions. 
Metrics are reported as mean $\pm$ standard deviation obtained from repeated stratified 5-fold cross-validation on the combined validation and test data.}
\label{tab:broach_ae_nested_cv}
\begin{tabularx}{\textwidth}{lRRRR}
\toprule
\textbf{Method} & \textbf{Precision} & \textbf{Recall} & \textbf{F1-Score} & \textbf{PR-AUC} \\
\midrule
AE-SVE-7 & 0.845 $\pm$ 0.100 & 0.876 $\pm$ 0.145 & 0.849 $\pm$ 0.101 & 0.926 $\pm$ 0.046 \\
AE-SVE-5 & 0.845 $\pm$ 0.100 & 0.889 $\pm$ 0.101 & 0.861 $\pm$ 0.071 & 0.907 $\pm$ 0.072 \\
Global MSE & 1.000 $\pm$ 0.000 & 0.507 $\pm$ 0.154 & 0.659 $\pm$ 0.144 & 0.643 $\pm$ 0.131 \\
Global MAE & 0.880 $\pm$ 0.146 & 0.462 $\pm$ 0.166 & 0.589 $\pm$ 0.161 & 0.539 $\pm$ 0.140 \\
\bottomrule
\end{tabularx}
\end{table}

\section{Benchmark Setup and Results} 
The benchmark evaluates the runtime performance of the complete anomaly detection pipeline, including spectrogram generation, 
CNN autoencoder inference, and anomaly scoring. All experiments were conducted using the same Docker image and dataset comprising 101 samples. 

\subsection{Evaluated Hardware Platforms} 

The benchmark was performed on four platforms: an Intel Core i7-12700H workstation (64 GB RAM), a Siemens Nanobox PC with an Intel Atom E3845 processor (8 GB RAM), 
a Raspberry Pi 4 (2 GB RAM), and a RevPi Connect+ based on a Raspberry Pi Compute Module 3 featuring a 1.2 GHz quad-core processor and 1 GB RAM.
\FloatBarrier

\subsection{Benchmark Results} 

\vspace{-6pt}

\begin{table}[H] 
\small
	\caption{Runtime performance of the complete inference pipeline. Reported latencies correspond to the mean execution time per sample over 101 benchmark runs in ms. 
	Throughput (TP) is given in samples/s.} 
	\label{tab:benchmark_results} 
	\begin{tabularx}{\textwidth}{lrRRRr} 
		\toprule \textbf{Platform} & \textbf{Prepro. [ms]} & \textbf{Infer. [ms]} & \textbf{Score [ms}] & \textbf{Total [ms]} & \textbf{TP [Samples/s]} \\ 
		\midrule Workstation & 128.7 & 4.9 & 0.4 & 134.0 & 7.5 \\ 
		Nanobox & 156.2 & 4.5 & 0.5 & 161.2 & 6.2 \\ 
		Rasp. Pi 4 & 704.1 & 44.4 & 1.7 & 750.2 & 1.3 \\
		RevPi Connect+ & 1472.5 & 49.6 & 4.7 & 1526.8 & 0.65 \\
		\bottomrule 
	\end{tabularx} 
\end{table}

\subsection{Discussion} 
The benchmark results indicate that the proposed deployment pipeline is portable across heterogeneous hardware architectures, 
ranging from industrial x86 edge devices to resource-constrained ARM-based systems. While overall runtime varies considerably 
between platforms, the relative contribution of preprocessing remains dominant. Consequently, further runtime improvements are 
expected to be more effective in the preprocessing stage than in the neural network inference stage.
\subsection{Sensitivity to Bottleneck Dimensionality} 
\label{app:turning_bottleneck}

A supplementary sensitivity analysis was conducted to investigate the influence of the autoencoder bottleneck 
dimensionality on anomaly-detection performance. This analysis was performed during model development and is 
reported separately from the final run-wise grouped five-fold evaluation presented in 
Table~\ref{tab:turning_method_comparison}. Bottleneck dimensions of 8, 16, 32, and 64 were evaluated using
 multiple reconstruction-error scoring strategies. The purpose of this analysis was to assess whether the 
 performance of the CNN-AE depends strongly on the latent representation size; the results are not used as 
 an additional estimate of the final model performance.

\begin{table}[H]
\small
\caption{Sensitivity of the CNN-AE performance to bottleneck dimensionality and reconstruction-error scoring strategy on the turning dataset.}\label{tab:turning_bottleneck_baseline_cv}
\begin{adjustwidth}{-\extralength}{0cm}
\begin{tabularx}{\fulllength}{llCCCC}

\toprule
\textbf{Method} & \textbf{Score} & \textbf{Precision} & \textbf{Recall} & \textbf{F1} & \textbf{PR-AUC} \\
\midrule
CNN-AE (BN = 8) & global\_mae & 0.835 $\pm$ 0.105 & 0.849 $\pm$ 0.099 & 0.837 $\pm$ 0.076 & 0.848 $\pm$ 0.123 \\
CNN-AE (BN = 8) & global\_mse & 0.613 $\pm$ 0.332 & 0.703 $\pm$ 0.361 & 0.645 $\pm$ 0.332 & 0.849 $\pm$ 0.119 \\
CNN-AE (BN = 8) & sve\_max & 0.785 $\pm$ 0.152 & 0.641 $\pm$ 0.369 & 0.600 $\pm$ 0.261 & 0.800 $\pm$ 0.126 \\
CNN-AE (BN = 8) & sve\_topk & 0.766 $\pm$ 0.166 & 0.661 $\pm$ 0.342 & 0.619 $\pm$ 0.213 & 0.825 $\pm$ 0.140 \\
\addlinespace
CNN-AE (BN = 16) & global\_mae & 0.638 $\pm$ 0.329 & 0.545 $\pm$ 0.372 & 0.534 $\pm$ 0.314 & 0.855 $\pm$ 0.127 \\
CNN-AE (BN = 16) & global\_mse & 0.597 $\pm$ 0.323 & 0.539 $\pm$ 0.350 & 0.520 $\pm$ 0.288 & 0.824 $\pm$ 0.127 \\
CNN-AE (BN = 16) & sve\_max & 0.509 $\pm$ 0.297 & 0.496 $\pm$ 0.316 & 0.487 $\pm$ 0.285 & 0.745 $\pm$ 0.144 \\
CNN-AE (BN = 16) & sve\_topk & 0.646 $\pm$ 0.127 & 0.589 $\pm$ 0.276 & 0.555 $\pm$ 0.162 & 0.778 $\pm$ 0.129 \\
\addlinespace
CNN-AE (BN = 32) & global\_mae & 0.734 $\pm$ 0.200 & 0.865 $\pm$ 0.096 & 0.777 $\pm$ 0.133 & 0.858 $\pm$ 0.129 \\
CNN-AE (BN = 32) & global\_mse & 0.658 $\pm$ 0.188 & 0.939 $\pm$ 0.052 & 0.753 $\pm$ 0.110 & 0.860 $\pm$ 0.120 \\
CNN-AE (BN = 32) & sve\_max & 0.721 $\pm$ 0.217 & 0.660 $\pm$ 0.332 & 0.637 $\pm$ 0.292 & 0.798 $\pm$ 0.138 \\
CNN-AE (BN = 32) & sve\_topk & 0.729 $\pm$ 0.199 & 0.699 $\pm$ 0.303 & 0.661 $\pm$ 0.248 & 0.817 $\pm$ 0.149 \\
\addlinespace
CNN-AE (BN = 64) & global\_mae & 0.691 $\pm$ 0.204 & 0.898 $\pm$ 0.090 & 0.760 $\pm$ 0.148 & 0.848 $\pm$ 0.129 \\
CNN-AE (BN = 64) & global\_mse & 0.659 $\pm$ 0.206 & 0.925 $\pm$ 0.104 & 0.748 $\pm$ 0.157 & 0.849 $\pm$ 0.118 \\
CNN-AE (BN = 64) & sve\_max & 0.540 $\pm$ 0.145 & 0.946 $\pm$ 0.076 & 0.675 $\pm$ 0.119 & 0.798 $\pm$ 0.137 \\
CNN-AE (BN = 64) & sve\_topk & 0.600 $\pm$ 0.183 & 0.951 $\pm$ 0.060 & 0.720 $\pm$ 0.137 & 0.832 $\pm$ 0.132 \\

\bottomrule
\end{tabularx}
\end{adjustwidth}
\end{table}

\subsection{Fold-Level Results for the Turning Dataset} 
\label{app:turning_cv}

To complement the aggregated results reported in Table~\ref{tab:turning_method_comparison}, 
Table~\ref{tab:turning_fold_results} reports the PR-AUC obtained on each held-out outer fold. 
The CNN-AE and classical baselines were evaluated using the same run-wise grouped five-fold 
assignments and the same group-wise calibration protocol. The fold-level results illustrate the 
variability between held-out machining runs underlying the mean and standard deviation reported in 
the main results.

\begin{table}[H] 
	 
	\caption{Fold-level PR-AUC values for the turning dataset under run-wise grouped five-fold cross-validation.} 
	\label{tab:turning_fold_results} 
\begin{tabular}{lcccc}
\toprule
\textbf{Outer Fold} & \textbf{CNN-AE (Global MAE)} & \textbf{Isolation Forest} & \textbf{One-Class SVM} & \textbf{PCA} \\
\midrule
1 & 1.0000 & 0.9914 & 0.9838 & 0.9098 \\
2 & 0.5170 & 0.6894 & 0.6962 & 0.6187 \\
3 & 0.8672 & 0.9421 & 0.6406 & 0.5198 \\
4 & 0.9758 & 1.0000 & 0.8788 & 0.9473 \\
5 & 0.7053 & 0.7870 & 0.5691 & 0.6223 \\
\bottomrule
\end{tabular}
\end{table}

\bibliographystyle{IEEEtran}
\bibliography{literatur}

\end{document}